\documentclass[final]{cvpr}

\usepackage{times}
\usepackage{epsfig}
\usepackage{graphicx}
\usepackage{amsmath}
\usepackage{amssymb}
\usepackage[utf8]{inputenc} % allow utf-8 input
\usepackage[T1]{fontenc}    % use 8-bit T1 fonts
\usepackage{hyperref}       % hyperlinks
\usepackage{url}            % simple URL typesetting
\usepackage{layouts}
\usepackage{makecell}
\usepackage{booktabs}       % professional-quality tables
\usepackage{amsfonts}       % blackboard math symbols
\usepackage{nicefrac}       % compact symbols for 1/2, etc.
\usepackage{microtype}      % microtypography
\usepackage{color}
\usepackage{xcolor}
\usepackage{xfrac}
\usepackage{multirow}
\usepackage{paralist}
\usepackage[subrefformat=parens]{subcaption}
\usepackage{microtype}
\usepackage{xspace}
\usepackage{amsmath,amsfonts,bm}
\usepackage{amssymb}
\usepackage{mathtools}
\usepackage{arydshln}
\usepackage{cuted}
\usepackage{capt-of}
\usepackage{multirow}
\usepackage{scalerel}

\usepackage{amsmath,amsfonts,bm,relsize,dsfont}

\newcommand{\myparagraph}[1]{\vspace{4pt} \noindent \textbf{#1}}

\newcommand{\notation}[1]{\ensuremath{#1}\xspace}

\def\Input{\notation{\bm{x}}} % input
\def\Output{\notation{\bm{y}}} % input
\def\Pixel{\notation{p}} % input
\def\LowresInput{\notation{\bm{x}_l}} % input
\def\TotalDownsamplingFactor{\notation{S}} % input
\def\DownsamplingFactor{\notation{S_1}} % input
\def\LearnedDownsamplingFactor{\notation{S_2}} % input
\def\LowresNet{\notation{G}}
\def\HighresNet{\notation{f}}
\def\HighresParams{\notation{\phi}}

\def\Re{\mathds{R}}

\definecolor{MyDarkBlue}{rgb}{0,0.08,1}
\definecolor{MyDarkGreen}{rgb}{0.02,0.6,0.02}
\definecolor{MyDarkRed}{rgb}{0.8,0.02,0.02}
\definecolor{MyDarkOrange}{rgb}{0.40,0.2,0.02}
\definecolor{MyPurple}{RGB}{111,0,255}
\definecolor{MyRed}{rgb}{1.0,0.0,0.0}
\definecolor{MyGold}{rgb}{0.75,0.6,0.12}
\definecolor{MyDarkgray}{rgb}{0.66, 0.66, 0.66}
\definecolor{MyDarkCyan}{rgb}{0.05, 0.55, 0.45}

\def\cvprPaperID{1509} % *** Enter the CVPR Paper ID here
\def\confYear{CVPR 2021}

\begin{document}

%%%%%%%%% TITLE
\title{Spatially-Adaptive Pixelwise Networks for Fast Image Translation}

\author{Tamar Rott Shaham$^1$ \quad
Micha\"{e}l Gharbi$^2$ \quad
Richard Zhang$^2$ \quad
Eli Shechtman$^2$ \quad
Tomer Michaeli$^1$ \\
\\
$^1$Technion \hspace{3cm}
$^2$Adobe Research\\
}

\maketitle

\begin{abstract}

We introduce a new generator architecture, aimed at fast and efficient
high-resolution image-to-image translation. We design the generator to be an
extremely lightweight function of the full-resolution image. In fact, we use
\textit{pixel-wise} networks; that is, each pixel is processed independently of
others, through a composition of simple affine transformations and
nonlinearities. We take three important %crucial 
steps to equip such a seemingly simple
function with adequate expressivity. First, the parameters of the pixel-wise
networks are \textit{spatially varying}, so they can represent a broader function
class than simple $1 \times 1$ convolutions. Second, these parameters are
\textit{predicted} by a fast convolutional network that processes an
aggressively low-resolution representation of the input. %this makes the model fully adaptive to the input image. 
Third, we augment the input image by concatenating a
sinusoidal encoding of spatial coordinates, which provides an effective inductive bias for generating realistic novel high-frequency image content. As a result, our model is up to $18\times$ faster than state-of-the-art baselines. %for images of size $1024 \times 1024$ %\todo{[$18\times$ comparing to CC-FPSE for $512 \times 512$. CC-FPSE can't be traind on $1024 \times 1024$]}. 
We achieve this speedup while generating comparable
visual quality across different image resolutions and translation domains.
% \keywords{}
%We would like to encourage you to list your keywords within the abstract section
\end{abstract}

\vspace{-0.5cm}
\section{Introduction}

\begin{figure}[t]
    \centering
    \includegraphics[height=9cm]{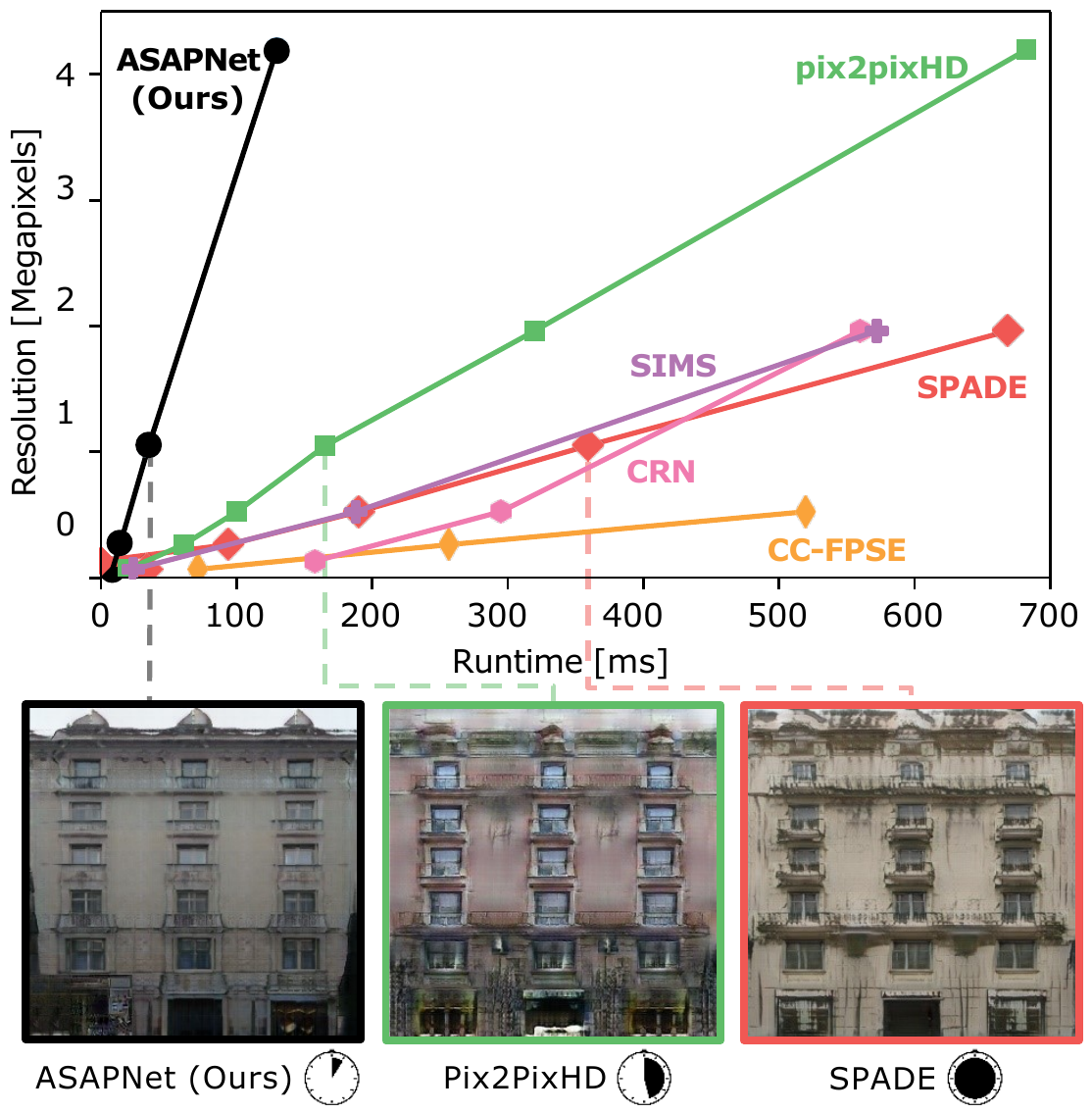}
    %\includegraphics[height=8cm]{figs/runtime_vs_imgsize_Mpix_v9_log.pdf}
    %\includegraphics[height=2cm]{figs/qualitative/facade05_label.png}
    %\includegraphics[height=2cm]{figs/qualitative/facade05_spade.png}
    %\includegraphics[height=2cm]{figs/qualitative/facade05_pix2pixHD.png}
    %\includegraphics[height=2cm]{figs/qualitative/facade05_asap.png}
    %\label{fig:run-time}
    %\vspace{-0.5cm}
    \caption{\textbf{Fast image %-to-image
    translation.} Our %Thanks to a 
    novel spatially adaptive pixelwise design %, our model can 
    enables generating high-resolution images at significantly lower runtimes than existing methods, while maintaining high visual quality. Particularly, as seen in the plot our model is $2$-$18\times$ faster than baselines \cite{liu2019learning,Park_2019_CVPR,qi2018semi,wang2018pix2pixHD,chen2017photographic}, depending on resolution.
    %our method (ASAPNet) is up to $18\times$ faster than CC-FPSE~\cite{liu2019learning}, $10\times$ faster than SPADE~\cite{Park_2019_CVPR}, SIMS~\cite{qi2018semi}, and CRN~\cite{chen2017photographic} and $6\times$ faster than pix2pixHD~\cite{wang2018pix2pixHD}
    %$8\times$ to $18\times$ faster than CC-FPSE~\cite{liu2019learning}, $4\times$ to $10\times$ faster than SPADE~\cite{Park_2019_CVPR} and $2\times$ to $6\times$ faster than pix2pixHD~\cite{wang2018pix2pixHD}, depending on resolution. %\todo{[add 3 images for a vertical cut (\eg. 1Mpix), and also refer to visual and quantitative result at section \ref{sec:qualitative}]} 
    %\tamar{i. linear or log scale? ii. isn't it weird that "ours" is leftmost?}
    }
    \vspace{-0.3cm}
    \label{fig:fig_0}
\end{figure}

\begin{figure*}
	\centering
	\captionsetup[subfigure]{labelformat=empty,justification=centering,aboveskip=1pt,belowskip=1pt}
	%%%%%%%%%%%%%%%%%%
	\begin{subfigure}[t]{0.19\textwidth}
	    \vspace{-.3mm}
		\centering
		\vspace{1\fontcharht\font`B}
		\caption{\footnotesize{$1024\times 1024$ label map}}
		\includegraphics[width=1\linewidth]{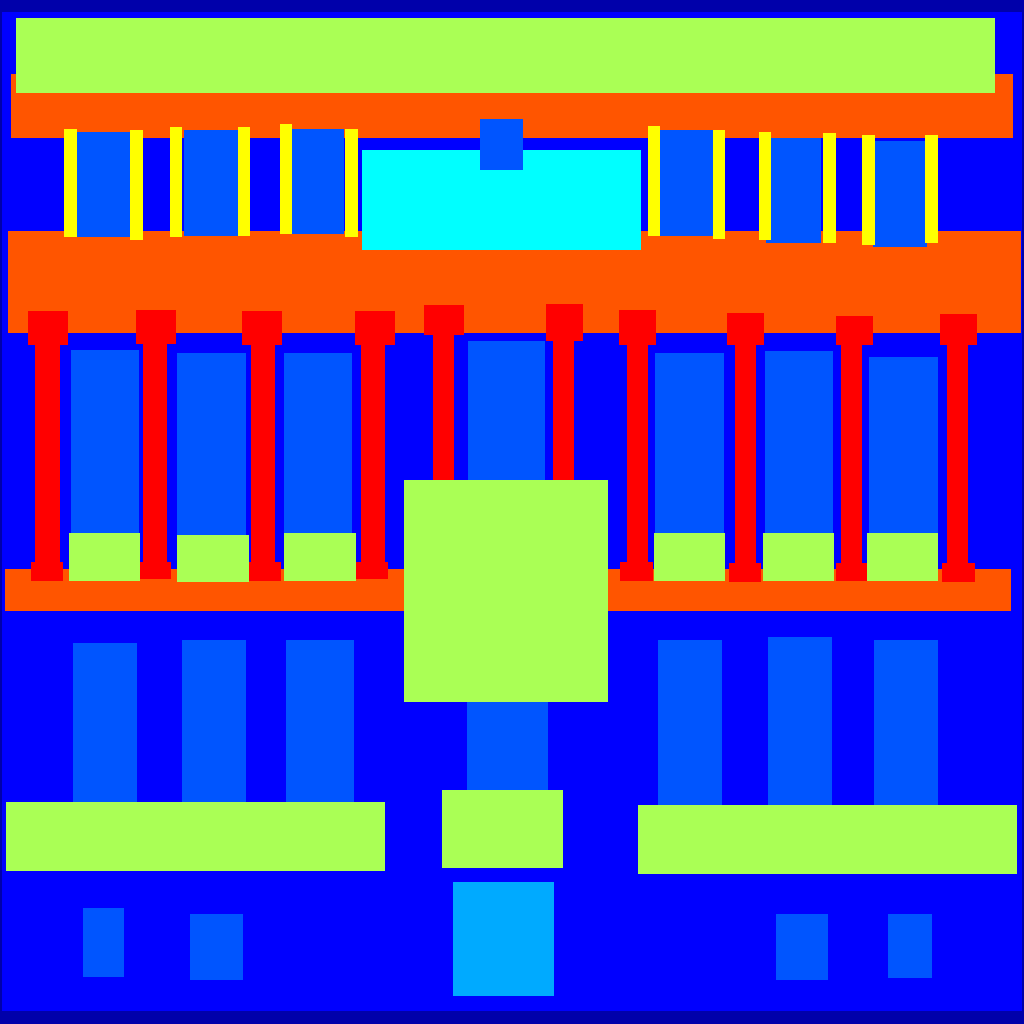}
	\end{subfigure}
	\begin{subfigure}[t]{0.19\textwidth}
	    \vspace{-.3mm}
		\centering
		\vspace{1\fontcharht\font`B}
		\caption{\footnotesize{ground truth}}
		\includegraphics[width=1\linewidth]{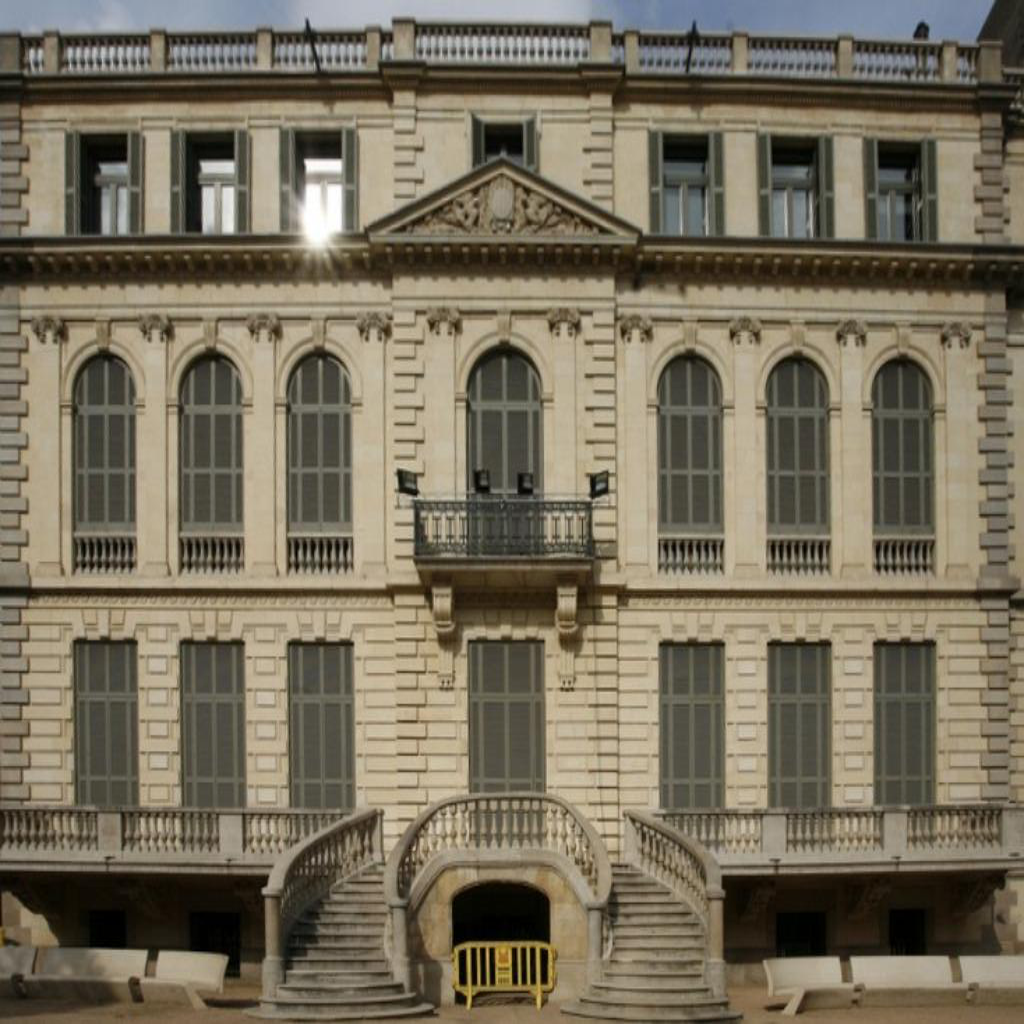}
		% \caption{vector sculpting}
	\end{subfigure}
	\begin{subfigure}[t]{0.19\textwidth}
		\centering
		\caption{\footnotesize{SPADE~\cite{Park_2019_CVPR}, \textbf{359ms}}
		\includegraphics[height=2.5\fontcharht\font`B]{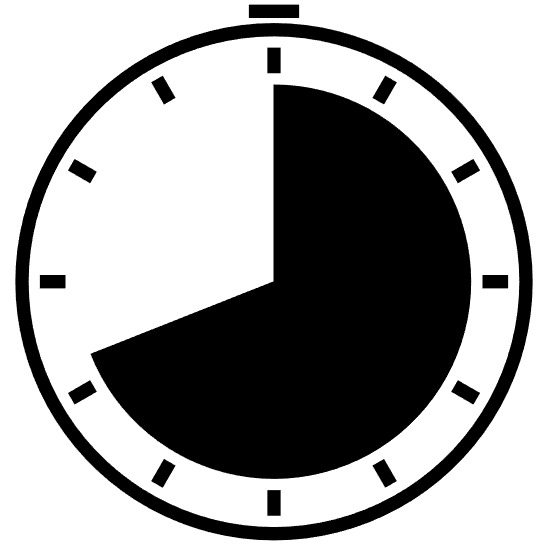}}
		\includegraphics[width=1\linewidth]{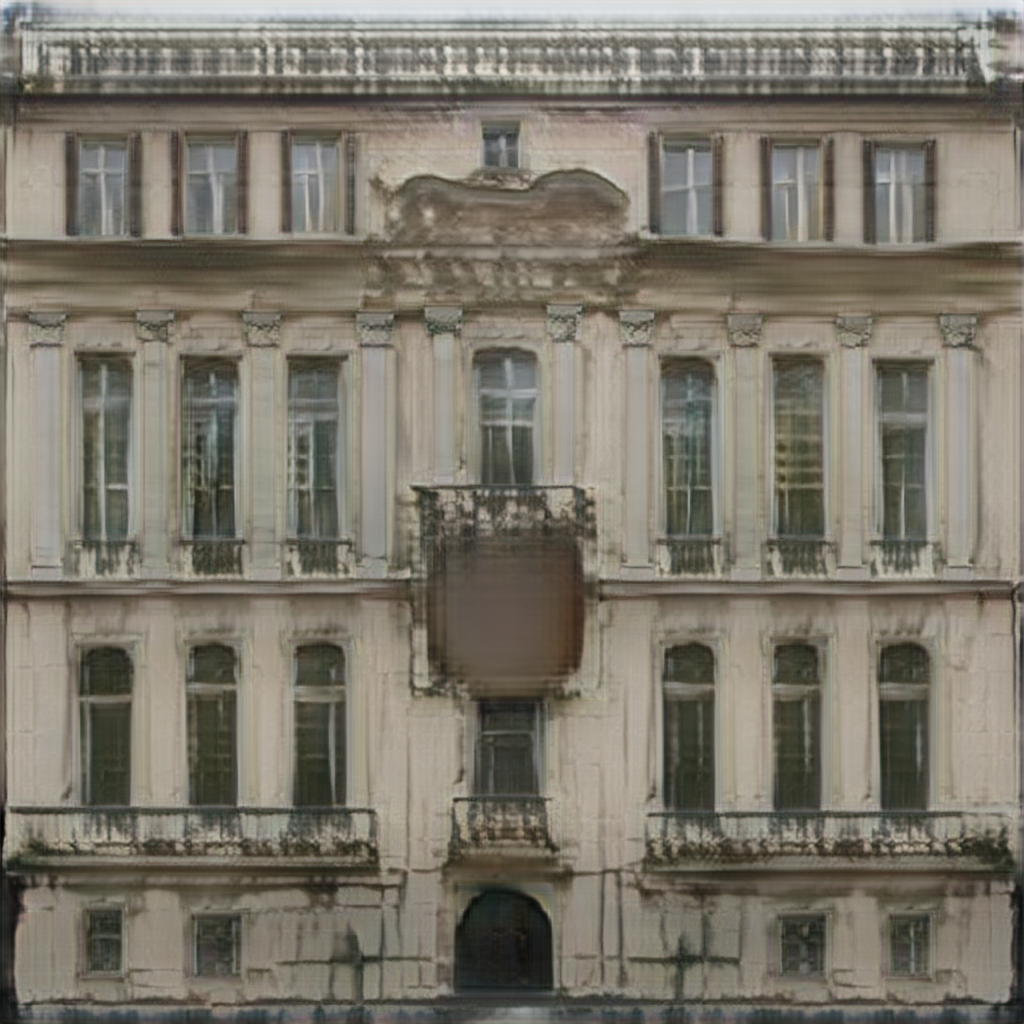}
		% \caption{vector sculpting}
	\end{subfigure}
	\begin{subfigure}[t]{0.19\textwidth}
		\centering
		\caption{
		\footnotesize{pix2pixHD~\cite{wang2018pix2pixHD}, \textbf{166ms}}
		\includegraphics[height=2.5\fontcharht\font`B]{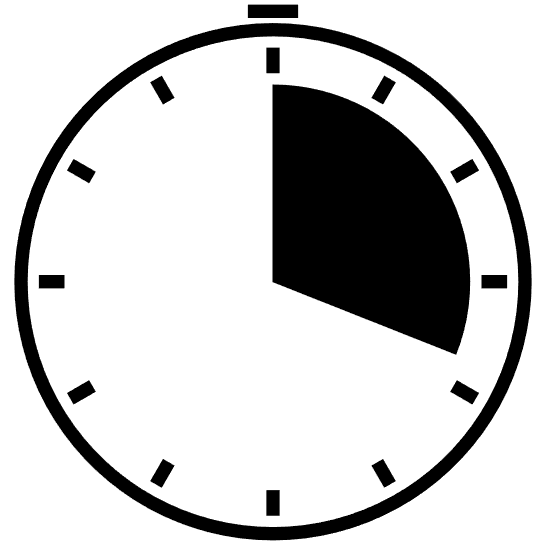}}
		\includegraphics[width=1\linewidth]{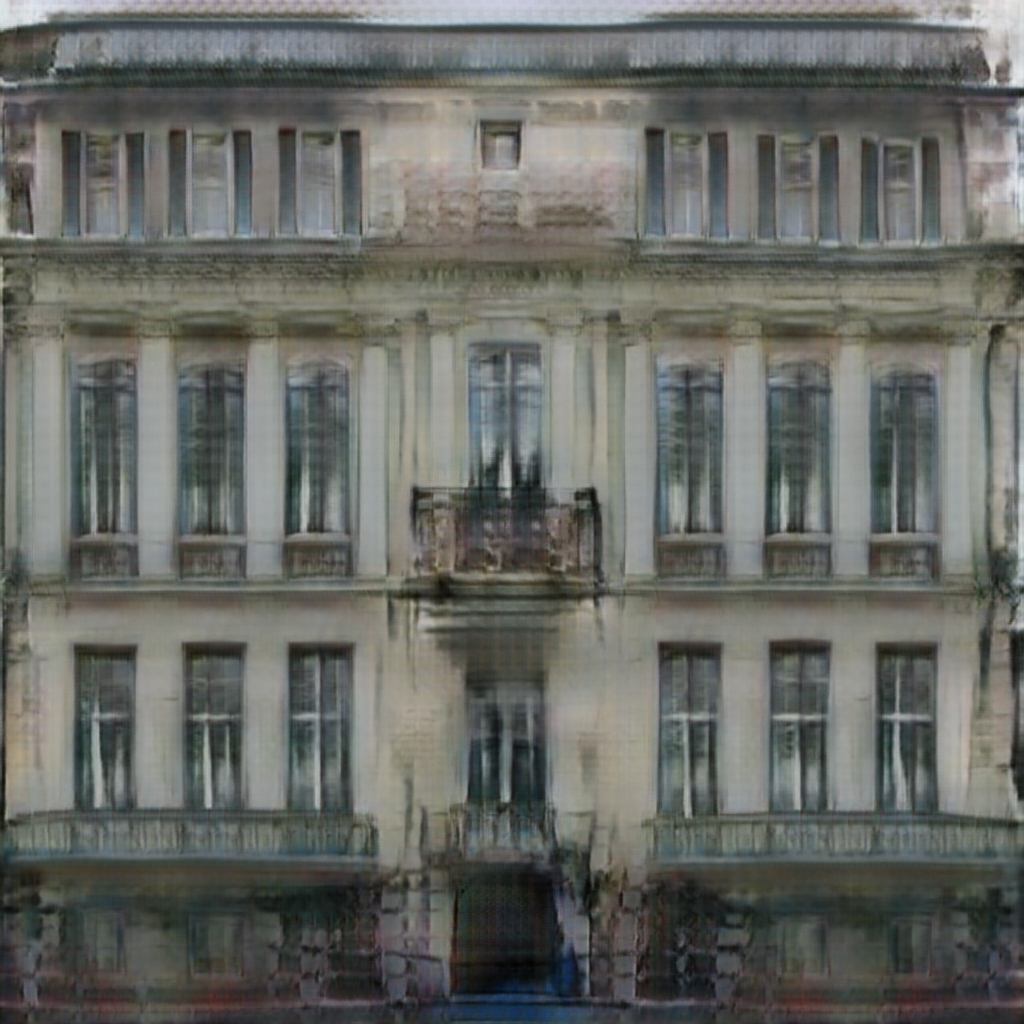}
		% \caption{vector sculpting}
	\end{subfigure}
	\begin{subfigure}[t]{0.19\textwidth}
		\centering
		\caption{\footnotesize{ASAP-Net (ours), \textbf{35ms}}
		\includegraphics[height=2.5\fontcharht\font`B]{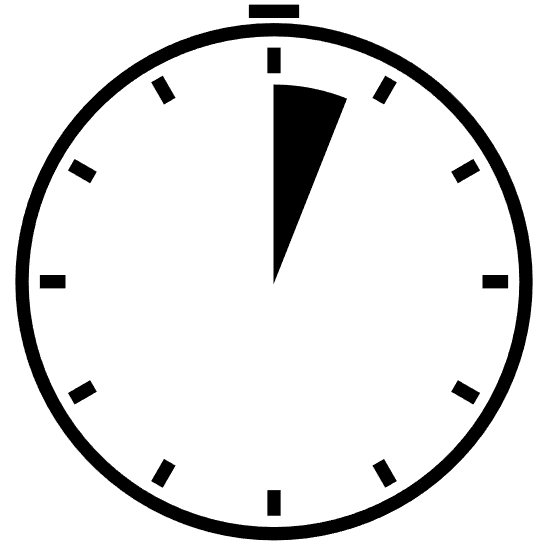}
		}
		\includegraphics[width=1\linewidth]{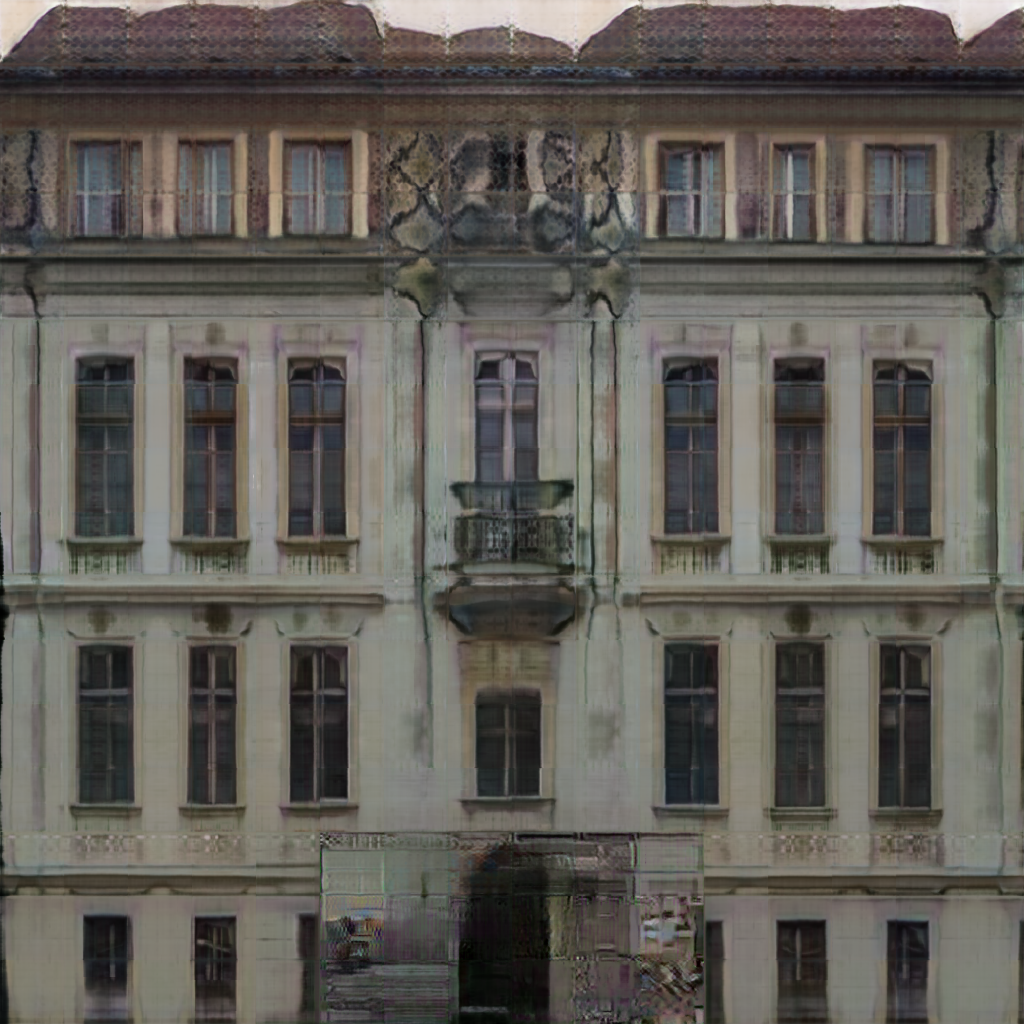}
		% \caption{vector sculpting}
	\end{subfigure}
	%%%%%%%%%%%%%%%
% 	\vspace{-1mm}
	\begin{subfigure}[t]{0.32\textwidth}
	    \vspace{1mm}
		\centering
		\vspace{0.88\fontcharht\font`B}
		\caption{\footnotesize{$512\times 1024$ label map}}
		\includegraphics[width=1\linewidth]{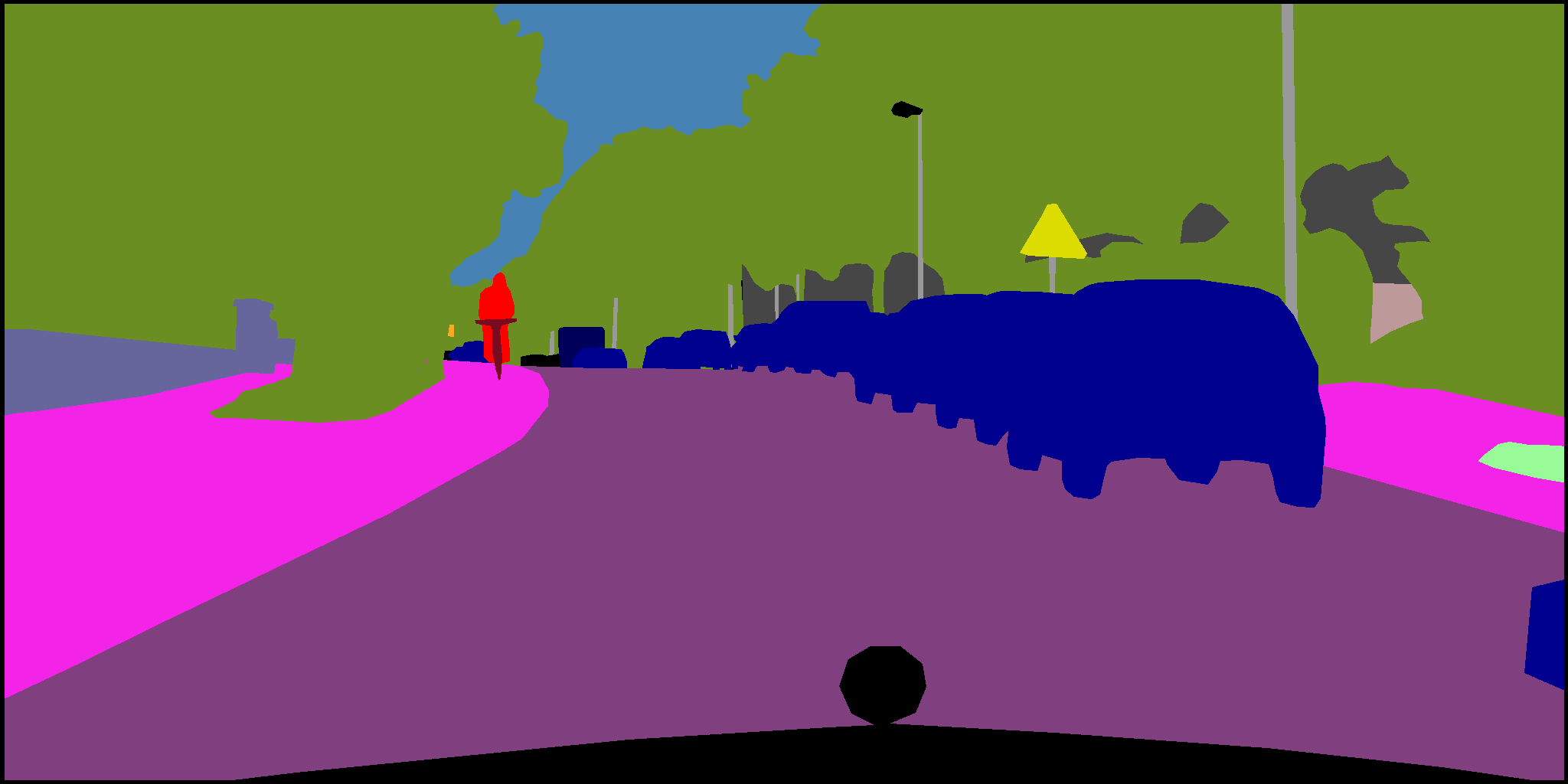}
	\end{subfigure}
	\begin{subfigure}[t]{0.32\textwidth}
    	\vspace{1mm}
		\centering
		\caption{\footnotesize{CC-FPSE~\cite{liu2019learning}, \textbf{520ms}}
		\includegraphics[height=2.5\fontcharht\font`B]{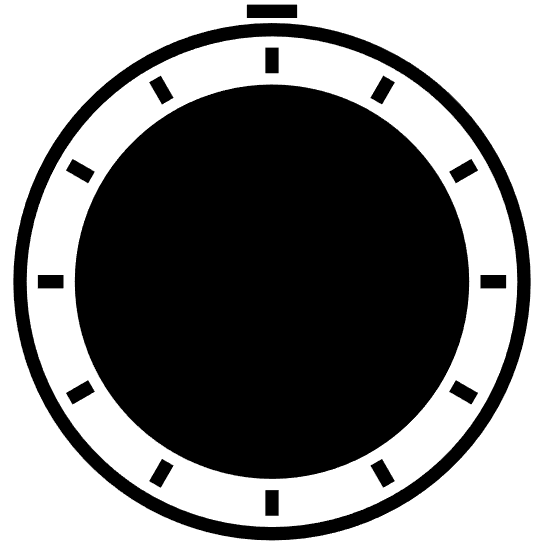}}
		\includegraphics[width=1\linewidth]{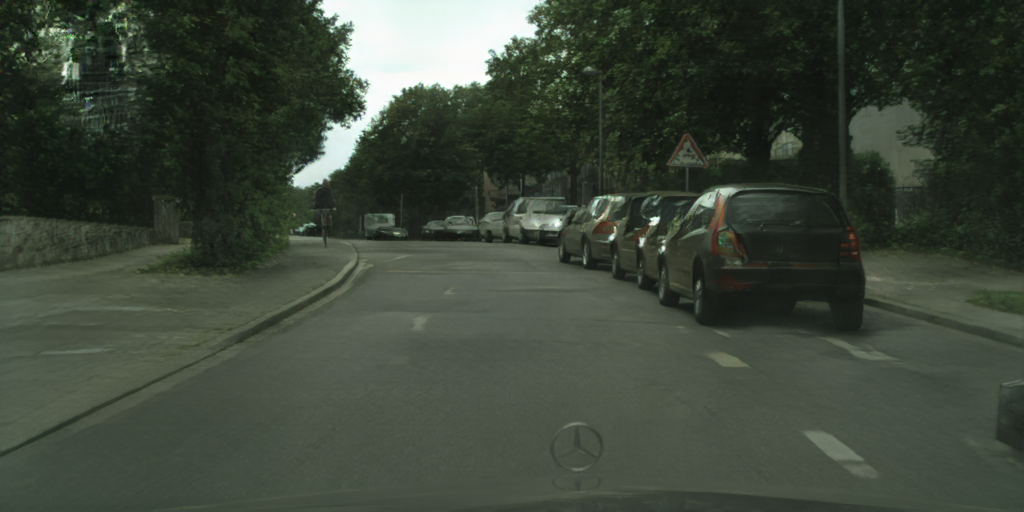}
	\end{subfigure}
	\begin{subfigure}[t]{0.32\textwidth}
	    \vspace{1mm}
		\centering
		\caption{\footnotesize{SPADE~\cite{Park_2019_CVPR}, \textbf{190ms}}
		\includegraphics[height=2.5\fontcharht\font`B]{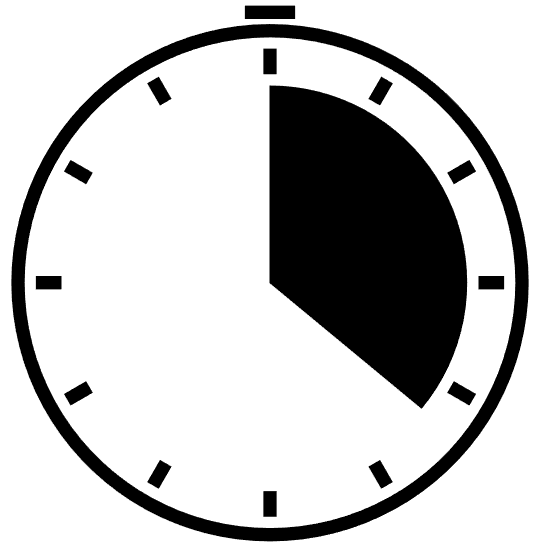}}
		\includegraphics[width=1\linewidth]{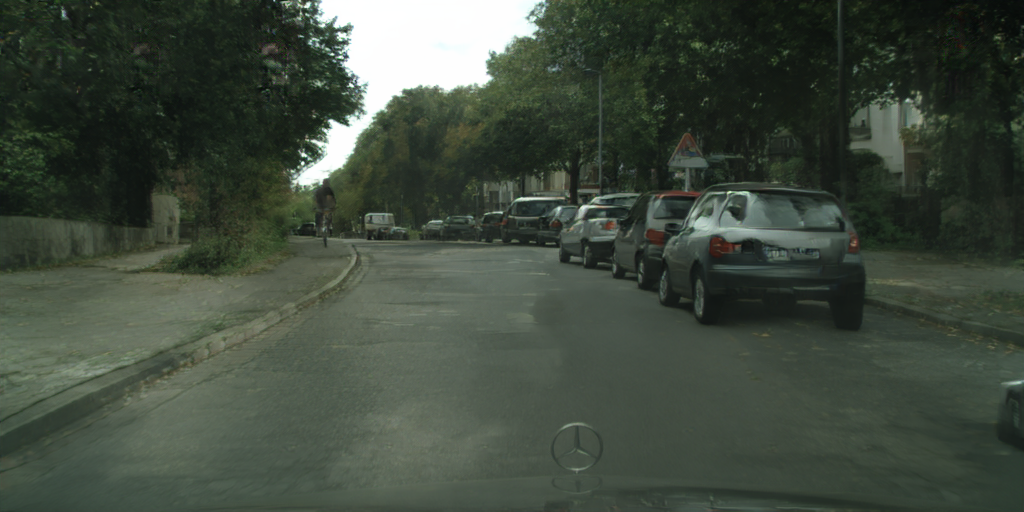}
	\end{subfigure}
	\par\smallskip
	\begin{subfigure}[t]{0.32\textwidth}
	    \vspace{-.5mm}
		\centering
		\vspace{0.88\fontcharht\font`B}
		\caption{\footnotesize{ground truth}}
		\includegraphics[width=1\linewidth]{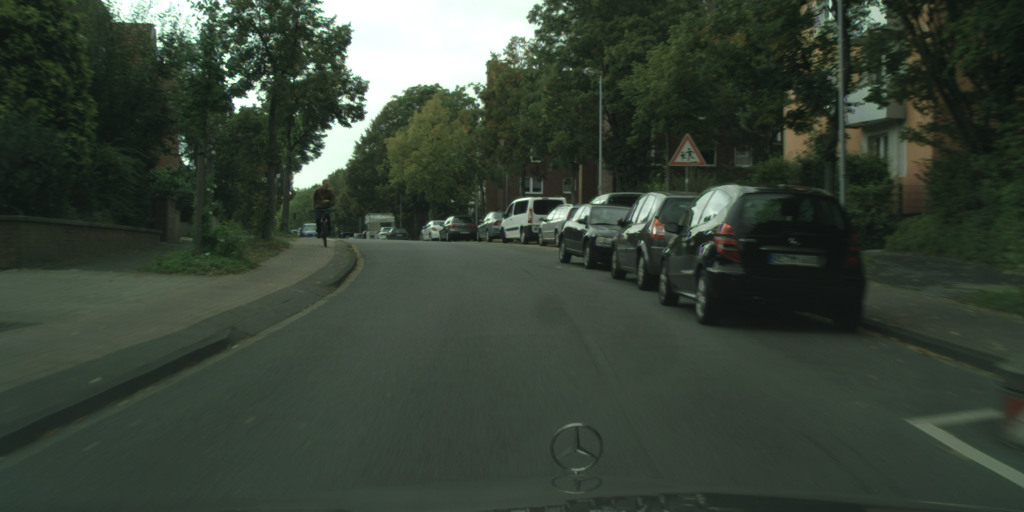}
	\end{subfigure}
	\begin{subfigure}[t]{0.32\textwidth}
	    \vspace{-.5mm}
		\centering
		\caption{\footnotesize{pix2pixHD~\cite{wang2018pix2pixHD}, \textbf{95ms}}
		\includegraphics[height=2.5\fontcharht\font`B]{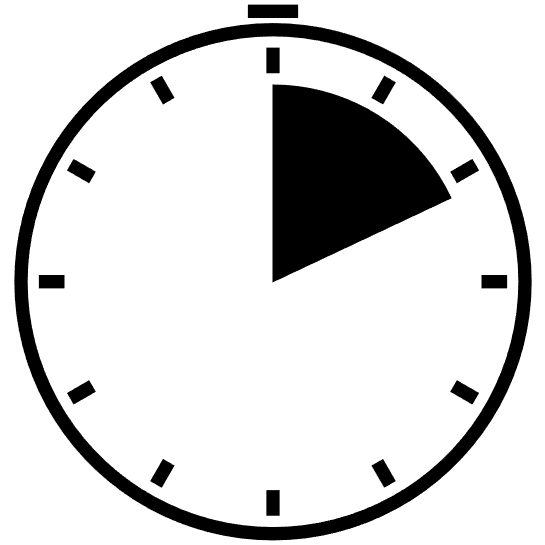}}
		\includegraphics[width=1\linewidth]{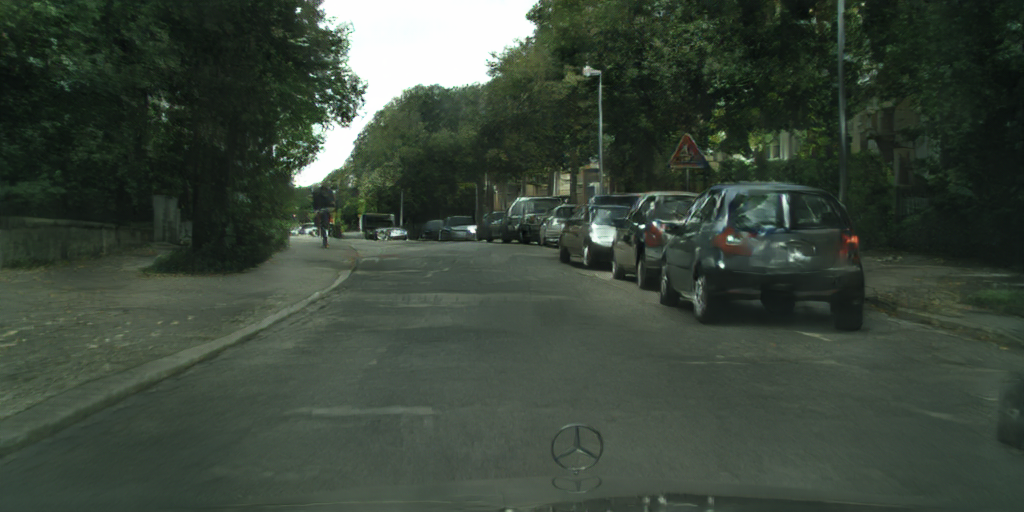}
	\end{subfigure}
	\begin{subfigure}[t]{0.32\textwidth}
	    \vspace{-.5mm}
		\centering
		\caption{\footnotesize{ASAP-Net (ours), \textbf{29ms}}
		\includegraphics[height=2.5\fontcharht\font`B]{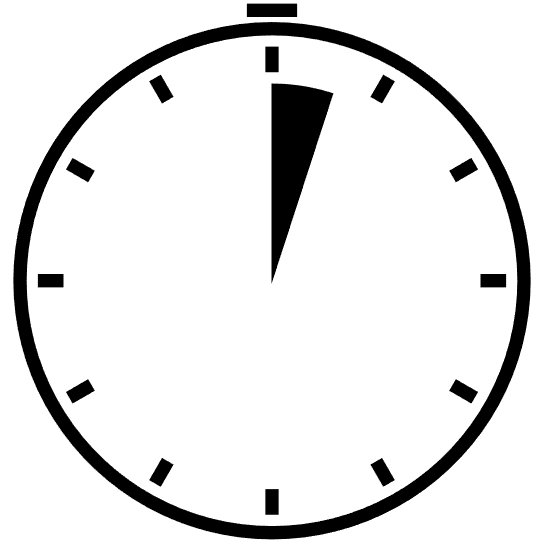}}
		\includegraphics[width=1\linewidth]{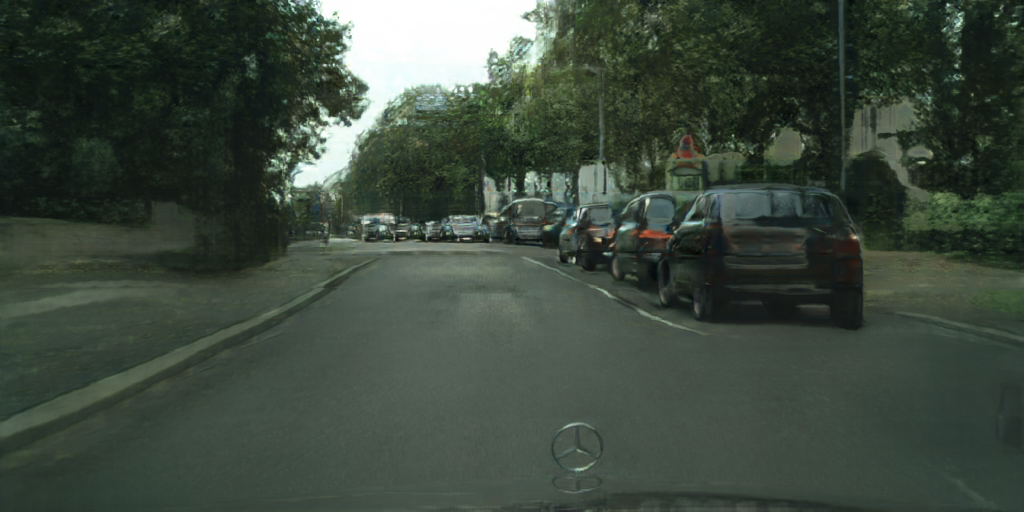}
	\end{subfigure}
	%%%%%%%%%%%%%%%
	%\setcounter{figure}{0}
	\caption{\label{fig:teaser}\textbf{Fast image-to-image translation.}
		%Thanks to a novel spatially adaptive pixelwise design,
	    Our model translates high-resolution images in short execution time comparing to baselines, while maintaining high visual quality.
	    In this example, we translate $512\times1024$ and a $1024\times1024$ %pixels 
	    label images into equal-sized outputs in only
	    $29$ and $35$ ms, respectively -- up to $10\times$ and $18\times$ faster
	    than the state-of-the-art methods %pix2pixHD~\cite{wang2018pix2pixHD}
	    %and 
	    SPADE~\cite{Park_2019_CVPR} and CC-FPSE~\cite{liu2019learning}, respectively  -- while maintaining the same visual
	    quality (see \S~\ref{sec:quantitative} for a user study and quantitative
		comparisons).
		Please note that CC-FPSE cannot be trained on $1024\times1024$ images due to memory constraints. %\tamar{what do you think about the timers visualiztion?}
	}
\end{figure*}{}

%Our network process each of the input pixels independently by a sequences of affine transformations and nonlinerities, which insure fast execution time. To gain expressivity, the parameters of these pixelwise networks are predicted from a drastically low-resolution representation of the input image, .         
%two streams architecture that exploits information from a low-resolution version of the input image and predicts the parameters of a  that acts on the full-resolution image. 
%Each of the image pixels is process independently by a spatially adaptive pixel-wise network.  

Translating images from one domain to another has been extensively studied
in recent years~\cite{isola2017image}.
Current approaches usually train a conditional Generative Adversarial Network
(GAN) to learn a direct mapping from one domain to the
other~\cite{zhu2017unpaired,chen2017photographic,wang2018pix2pixHD,Park_2019_CVPR,liu2019learning}.
Although these approaches have made rapid progress in terms of visual quality,
model size and inference time have also grown significantly.
The computational cost of these algorithms becomes even more acute when
operating on high resolution images, which is the most desirable setting in
typical real-world applications. 

In this work, we present a novel architecture, designed for fast image-to-image
translation\footnote{\url{https://tamarott.github.io/ASAPNet_web}}.
The key ingredient for an efficient runtime is a new generator that operates
\textit{pixelwise}: each pixel is processed independently from the others using
a pixel-specific, lightweight Multi-Layer Perceptron (MLP).
At first glance, the representation power of this generator should appear limited.
However, three key components make our network fully expressive.
First, in contrast to traditional convolutional networks, where network
parameters are shared across spatial positions, the parameters of the MLPs
\textit{vary spatially} so each pixel is effectively transformed by a different
function.
Second, the spatially-varying parameters are predicted at \text{low-resolution}
by a convolutional network that processes a drastically downsampled
representation of the input.
This makes the MLPs adaptive to the input image (i.e., the pixel-functions
depend on the input image itself).
Third, in addition to the input pixel values, the local MLPs consume a
sinusoidal encoding of the pixel's spatial position~\cite{vaswani2017attention}.
Together, these three components enable realistic output synthesis with
coherent and detailed image structures.

As a result, our model, which we coin ASAP-Net (A Spatially-Adaptive Pixelwise
Network), generates images of high visual quality at very short execution times,
even for high-resolution inputs (see Fig.~\ref{fig:fig_0}).
A visual comparison to state-of-the-art methods can be seen in Fig.~\ref{fig:teaser}. Here, our model processes $512\times1024$ and
$1024\times1024$ label maps in only 29 and 35 milliseconds on a GPU. 
This is over $4.5\times$, $10\times$ and
$18\times$ faster than the high-performance pix2pixHD~\cite{wang2018pix2pixHD}, SPADE~\cite{Park_2019_CVPR} and CC-FPSE~\cite{liu2019learning}
models, respectively, on the same hardware. 

We evaluate our model on several image domains and at various resolutions,
including transferring facade labels to building images and generating realistic
city scenery from semantic segmentation maps. Additionally, we show the flexibility of our method by testing on a markedly different translation task, predicting depth maps from indoor images. % a setting where existing translation methods such as SPADE struggles. % Additionally, we illustrate how, as opposed to methods like SPADE, our model can go beyond label maps and translate e.g.~indoor images to depth-maps.
In all cases, our model's runtime is considerably lower than existing state-of-the-art methods, while its visual quality is comparable. We confirm this with human perceptual studies, as well as by automatic metrics.

% old stuff

\section{Related Work}

\myparagraph{Convolutional image-to-image translation.}
Classic networks for image-to-image translation typically adopt a
sequential convolutional design, such as encoder-decoder based
models~\cite{isola2017image,ronneberger2015u,qi2018semi} or coarse-to-fine
strategies~\cite{chen2017photographic}.
Architectural improvements such as residual blocks~\cite{zhu2017unpaired}, conditional convolution blocks~\cite{liu2019learning} and
normalization
techniques~\cite{liu2017unsupervised,huang2018multimodal,Park_2019_CVPR} have
rapidly improved the output image quality.
Unfortunately, these improvements have also led to
models with higher complexity, increased computational costs and slower
runtime~\cite{li2020gan}, even with faster hardware.
Our work strives to break away from the purely sequential convolutional network
design: we synthesize full-resolution pixels using lightweight, spatially
adaptive pixel-wise networks, whose parameters are predicted by a fast
convolutional network running at a much lower resolution.

\myparagraph{Adaptive neural networks.}
Networks that can tailor their parameters to specific inputs have been used in
various contexts.
Co-occurence networks~\cite{shevlev2019coocurrence} can tune weights based on pixel co-occurences within a fixed-size window. 
Kernel-predicting
networks~\cite{bako2017kernel,mildenhall2018burst,xia2019basis} produce
spatially-varying linear image-filtering kernels.
Deformable convolutions~\cite{dai2017deformable} enable irregularly-shaped local
filters.
\cite{shelhamer2019blurring} proposed a model that rescales the
receptive field of local filters based on image content.
Adaptive normalization techniques~\cite{huang2017adaptive,Park_2019_CVPR} and activation techniques~\cite{qian2018adaptive, kligvasser2018xunit} modulate the parameters of these layers according to an input signal
(e.g., label maps).
Closer to our approach, hypernetworks~\cite{ha2016hypernetworks} have been used
in visual reasoning and question-answering~\cite{perez2018film}, video
prediction~\cite{jia2016dynamic}, one-shot
learning~\cite{bertinetto2016oneshot} and parametric image
processing~\cite{fan2018decouple}.
Our low-resolution convolutional network is an instance of a hypernetwork.
It predicts spatially-varying parameters for a family of pixel-wise fully
connected networks. 
Hypernetwork functional image representations~\cite{klocek2019hyperfunctional}
are similar but use a single fully-connected network to encode
the image globally and have been used as image autoencoders. 
Our pixelwise networks are local, optimized for speed, and used in a
generative context.

\myparagraph{Trainable efficient image transformations.}
The key to our model's efficiency is that most of the computationally heavy
inference is performed at extreme low-resolution, while the high-resolution
synthesis is comparatively lightweight.
This strategy has been employed in the past to efficiently approximate costly
image filters~\cite{gharbi2015transform,chen2016bilateral}.
%
% Permutohedral lattice
%
The work most closely related to ours in this context is that of~\cite{gharbi2017deep}, where a network predicts a set of affine color
transformations from a low-resolution image, which are then upsampled to the full image size
in an edge-aware fashion%using a bilateral grid
~\cite{chen2007real}.
The local affine color transformations model is a good fit for photographic
edits (e.g., tone mapping or Local Laplacian filtering~\cite{paris2011local}), but
not expressive enough for more complex image-to-image translation tasks.
In particular, this model purely relies on existing edges in the input and
cannot synthesize new edges.

% DCT, Wavelets, continuous network based, image bases.
\myparagraph{Functional image representations.}
Another representation that inspired our design are Compositional Pattern
Producing Networks (CPPNs)~\cite{stanley2007cppn}.
A CPPN is a continuous functional mapping from spatial $(x,y)$ coordinates to
image colors.
Our pixelwise networks, which operate at full resolution, can be viewed as
spatially varying CPPNs.
Instead of the raw $(x,y)$ values, we encode pixel positions using sinusoids at various frequencies.
This is similar in spirit to Fourier CPPNs~\cite{tesfaldet2019fourierCppn},
although we do not perform any Fourier transform.
Similar %functional 
representations have been used to encode 3D
scenes~\cite{sitzmann2019scene,mildenhall2020nerf}.

\myparagraph{Network optimization and compression.}
%
% \rz{List NAS and compression work}
There exist many techniques to accelerate neural networks,
either by Neural Architecture Search~\cite{zoph2016nas,li2020gan},
weight-prunining~\cite{luo2017thinet}, parameter
quantization~\cite{zhou2017incremental,han2015deep} or low-rank approximation of
the filters~\cite{lebedev2014speeding}.
Our approach is largely orthogonal to these techniques. Our low-resolution
convolutional network can benefit from the same accelerations and our
full-resolution pixelwise operators are lightweight and highly parallelizable, and thus efficient by design.

% old stuff below

\section{Method}

\begin{figure*}[thp]
 \centering
\includegraphics[width=1.\hsize]{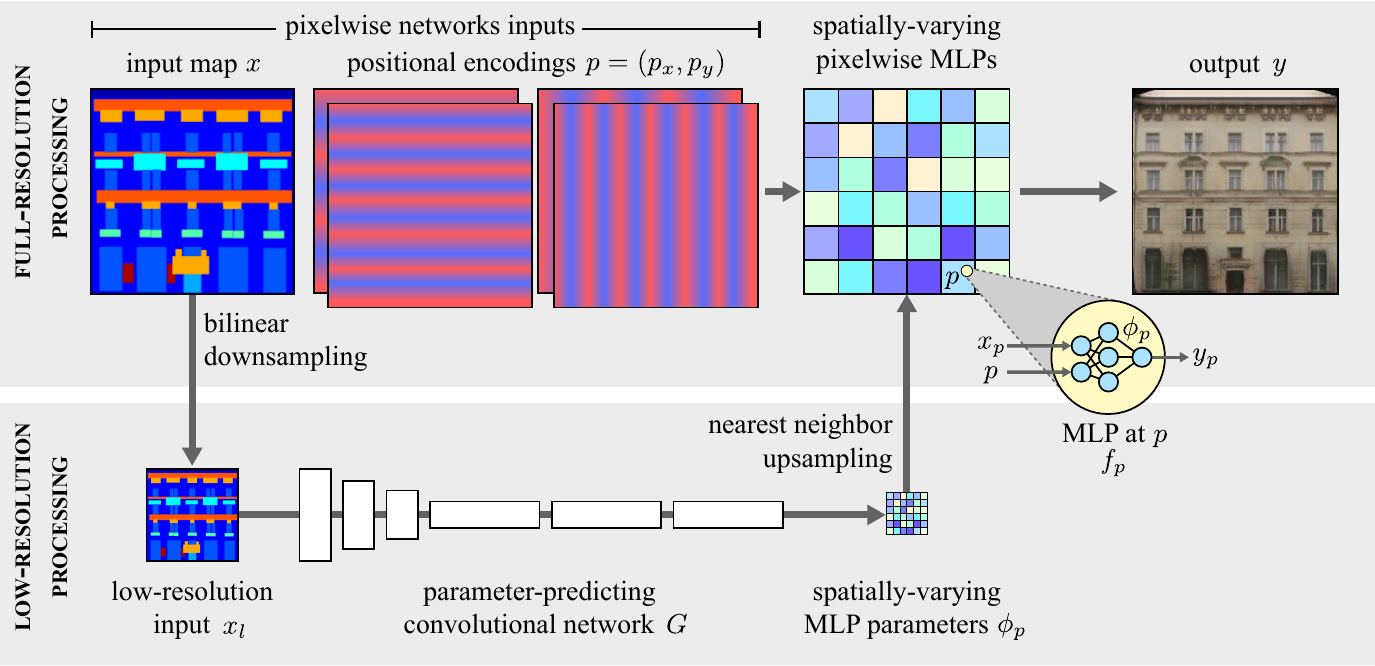}
\caption{\label{fig:arch}{\bf Architecture overview.} Our model first processes
   the input at very low-resolution, \LowresInput, to produce a tensor of weights and
   biases $\HighresParams_\Pixel$. These are upsampled back to full-resolution,
   where they parameterize pixelwise, spatially-varying MLPs
   $\HighresNet_\Pixel$ that compute the final output \Output from the
   high-resolution input \Input. In practice, we found nearest neighbor
   upsampling to be sufficient, which means that the MLP parameters are in fact
   piecewise constant.
} 
\end{figure*}

We consider image-to-image translation tasks, in which a neural network is
trained to implement a mapping between two image domains. That is, given a
high-resolution input image $\Input \in \Re^{H\times W\times C}$ from the source
domain, the output $\Output \in \Re^{H\times
W\times 3}$ should look like it belongs to the target domain.
In cases where the input is a label map, $C$ is the number of semantic classes
and $\Input$ is their one-hot representation. When the input is an image, $C=3$. 
Our goal is to design a network that is more efficient than
existing approaches yet produces the same high-fidelity outputs.

Inspired by HDRNet~\cite{gharbi2017deep}, we argue that although \emph{analyzing} and
understanding the image's content is a complex task and calls for an accordingly
deep and expressive neural network, this analysis need not be performed at
full-resolution. Furthermore, \emph{synthesizing} the output pixels should not require tens of convolutional layers.
Our strategy is thus twofold: first, we synthesize the high-resolution pixels
using lightweight and highly parallelizable operators (\S~\ref{sec:pixelwise}),
and second we perform the more costly image analysis at a very coarse resolution
(\S~\ref{sec:lowres}).
This design allows us to keep the full-resolution computation to a minimum,
perform the incompressible heavy computation on a much smaller input, while
maintaining a high output fidelity.
Our overall architecture is illustrated in Fig.~\ref{fig:arch}.

\subsection{Spatially Adaptive Pixelwise Networks}\label{sec:pixelwise}

At the full-resolution of the input image, computation comes at a premium.
To keep the runtime low, we model the output as a \emph{spatially-varying}
pointwise nonlinear transformation $\HighresNet_\Pixel$ of the high-resolution
input, where \Pixel denotes the pixel position. This is shown in the top branch in Fig.~\ref{fig:arch}.
%
% \begin{equation}
%    \Output_\Pixel = \HighresNet_\Pixel(\Input_\Pixel).
%    \label{eq:pixelwise}
% \end{equation}
%
% Pointwise -> fast and parallelizable, but also can be dangerous
The pointwise property makes the computation %trivially 
parallelizable, since each
pixel is now independent of the others. But without a careful design and a proper
choice of inputs, the lack of spatial context can also limit the model's expressiveness.

% Spatially varying + Adaptive
To preserve spatial information, we use two mechanisms. First, each pixelwise
function $\HighresNet_\Pixel$ takes as input the pixel's coordinates \Pixel, in
addition to its color value $\Input_\Pixel$. Second, the pixelwise functions are
parameterized with spatially-varying parameters $\HighresParams_\Pixel$ and
conditioned on the input image~\Input.
% \LowresInput, an aggressively downsampled version of the
%
Specifically, each $\HighresNet_\Pixel$ is a Multi-Layer Perceptron (MLP) with
ReLU activations defining the mapping
\begin{equation}
   % \HighresNet_\Pixel : (\Input_\Pixel, \Pixel)  \mapsto
   % \HighresNet(\Input_\Pixel, \Pixel, ;\HighresParams_\Pixel) 
   \HighresNet_\Pixel(\Input_\Pixel, \Pixel)= \HighresNet(\Input_\Pixel, \Pixel,
   ;\HighresParams_\Pixel) \eqqcolon \Output_\Pixel,
   \label{eq:parametric_highres}
\end{equation}
where \HighresNet denotes the MLP architecture --- shared by all pixelwise functions
--- and $\HighresParams_\Pixel$ are the (spatially-varying)
weights and biases for the MLPs.
We found that MLPs with 5 layers and 64 channels per layer provide fast
execution and good visual quality. 

Under this model, different input images yield different sets of per-pixel
parameters. This in turn means each image is processed by completely
different pointwise networks that vary across space.
Without these two properties --- input-adaptive and spatially-varying parameters
--- the pointwise approach would reduce to a much less expressive model: a
static stack of $1\times1$ convolutional layers.
Figure~\ref{fig:abl:constant} shows the results of using this 
spatially-uniform alternative.

\subsection{Predicting pixelwise network parameters from a low-resolution input}\label{sec:lowres}

For the transformation to be adaptive, we want $\HighresParams_\Pixel$
to be a function of the input image.
Predicting a (potentially large) parameter vector $\HighresParams_\Pixel$
for each pixel independently would be prohibitive.
Instead, the parameter vectors are predicted by a convolutional network
\LowresNet, operating on a much lower resolution image \LowresInput.
Specifically, the network \LowresNet outputs a grid of parameters at $\TotalDownsamplingFactor \times $ smaller resolution than $\Input$. This grid is then upsampled  using nearest neighbor interpolation
to obtain a parameter vector $\HighresParams_\Pixel$ for every high-resolution pixel~\Pixel, so that
\begin{equation}
   \HighresParams_\Pixel = \left[\LowresNet(\LowresInput)\right]_{\lfloor\frac{\Pixel}{\TotalDownsamplingFactor}\rfloor}.
   \label{eq:highres_params}
\end{equation}
For this step, our framework can in fact accommodate any upsampling scheme. However, empirically, we found that nearest-neighbor upsampling yields sufficiently good results, with the advantage of not requiring additional arithmetic operations for interpolation.

As seen in the bottom branch of Fig.~\ref{fig:arch}, the low-resolution computation starts with bilinearly downsampling
\Input by a factor \DownsamplingFactor to obtain
\LowresInput.
%That is, the size of the low-resolution image is fixed regardless of the full-resolution input dimensions.
%
This image is then processed by \LowresNet, which further reduces the spatial
dimensions by a factor \LearnedDownsamplingFactor using a sequence of strided-convolution layers.
The final output of the network \LowresNet is a tensor %$|\HighresParams_\Pixel|$
with channels corresponding to the weights and biases of the full-resolution pixelwise
networks.

We use $\LearnedDownsamplingFactor=16$
throughout and set \DownsamplingFactor so that \LowresInput 
has at most 256 pixels on the short axis of the image.
Therefore, the total downsampling $\TotalDownsamplingFactor\coloneqq\DownsamplingFactor\times\LearnedDownsamplingFactor$ depends
on the image size (e.g.~\TotalDownsamplingFactor $=64$ for an input of size $1024 \times 1024$). This means that our low-resolution stream processes a very low resolution representation of the input image (e.g.,~only $16 \times 16$ pixels for images with aspect ratio of~1), which dramatically reduces the
computation cost of \LowresNet. 
Additional network details can be found in the \href{https://tamarott.github.io/ASAPNet_web/resources/SM.pdf}{supplemental material}.

\begin{figure}[t]
    \centering
    %\begin{subfigure}[]{0.49\textwidth}
    %\includegraphics[height=5.2cm]{figs/runtime_vs_imgsize_Mpix_v6.pdf}
    %\caption{Runtime vs. image size}\label{fig:run-time}
    %\end{subfigure}
    %\begin{subfigure}[]{0.49\textwidth}
    \includegraphics[height=5.2cm]{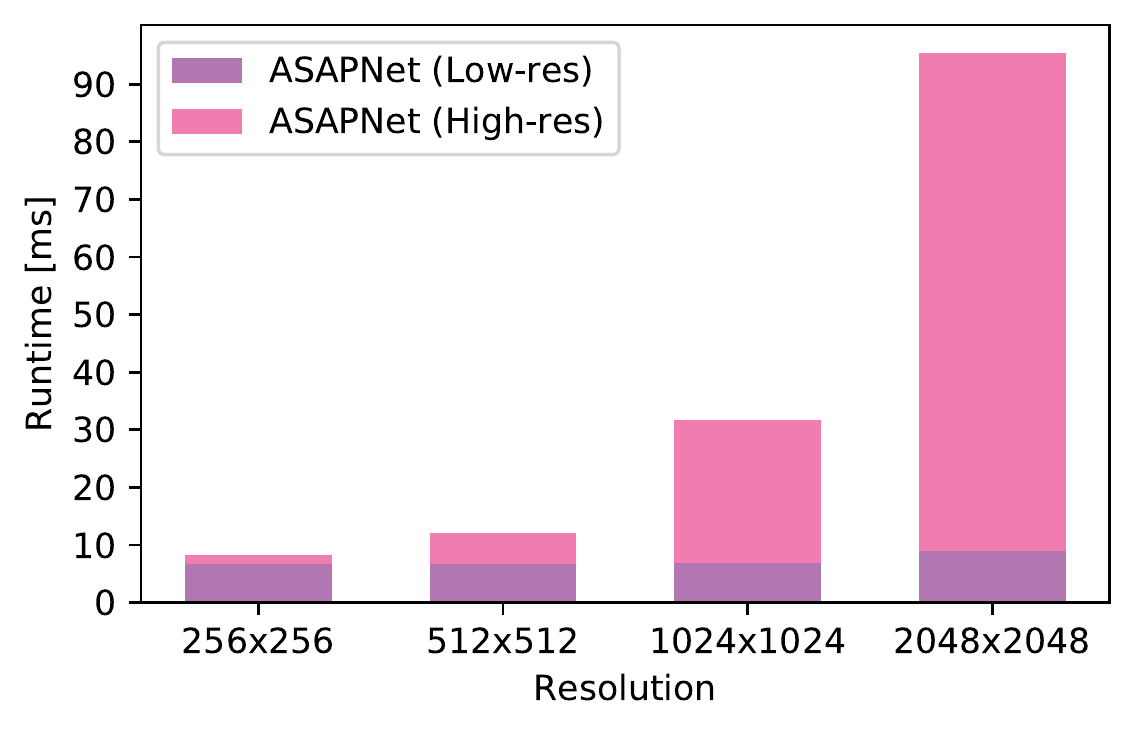}
    %\caption{HR vs. LR stream runtime}\label{fig:LRvsHR}
    %\end{subfigure}
    \caption{\label{fig:LRvsHR}%Our method (ASAP) is 8-18 $\times$ faster than CC-FPSE~\cite{liu2019learning}, 4-10$\times$ faster than
    %SPADE~\cite{Park_2019_CVPR} and 2-6$\times$ faster than
    %pix2pixHD~\cite{wang2018pix2pixHD}, depending on the resolution. 
    We show a breakdown of our method's runtime between the low-resolution convolutional network, and the full-resolution pixel-wise computation. Note that the runtime for the convolutional component remains roughly constant with resolution, while the pixelwise part scales linearly with the number of pixels to process (see SM for 4Mpix results).}
    %\vspace{-0.3cm}
\end{figure}{}

\subsection{Synthesizing high-resolution details using positional encodings}

Image translation requires hallucinating fine details, often finer 
than the input itself (e.g.~in the case of translating label
maps to images). However, our function parameters $\HighresParams_\Pixel$ are
predicted at low-resolution and then upsampled. This should somewhat limit their
ability to synthesize high frequency details.
We circumvent this limitation by augmenting the pixelwise functions
$\HighresNet_\Pixel$ to take an encoding of pixel position~\Pixel as additional
input~\cite{liu2018coordconv}.
Thus, like Compositional Pattern-Producing Networks (CPPN)~\cite{stanley2007cppn}, our
spatially-varying MLPs can learn to generate details finer than the sampling resolution of their parameters.
Furthermore, neural networks have been shown to have a spectral bias towards learning low-frequency signals first~\cite{rahaman2018spectral,ronen2019convergence}.
So, instead of passing~\Pixel directly to the MLP, as suggested by Equation~\eqref{eq:parametric_highres}, we found it useful to encode each component of the 2D pixel position $\Pixel=(\Pixel_x,\Pixel_y)$ as a vector of
sinusoids with frequencies higher than the upsampling factor.
This approach is similar in spirit to Fourier CPPN~\cite{tesfaldet2019fourierCppn}.
Specifically, in addition to the pixel value $\Input_\Pixel$, each MLP 
consumes $2\times2\times k$ additional input channels: 
$\big(\sin(2\pi \Pixel_x / 2^k), \cos(2 \pi \Pixel_x / 2^k)\big)$ for $k
=1,\ldots,\log_2(\TotalDownsamplingFactor)$ and similarly for~$\Pixel_y$.
The pixelwise MLPs can take advantage of these additional inputs to synthesize high frequency image patterns.
This encoding is shown in Figure~\ref{fig:arch}.
Figure~\ref{fig:abl:pos_encoding} illustrates their crucial role.

\subsection{Training and implementation details}\label{sec:optim}

Our contribution is our architectural design, and we follow best practices for the optimization procedure. We train our generator model adversarially with a multi-scale patch discriminator, as suggested by pix2pixHD~\cite{wang2018pix2pixHD}.
We follow the optimization procedure of SPADE~\cite{Park_2019_CVPR} which includes an adversarial hinge-loss~\cite{wang2017magan}, a perceptual
loss~\cite{dosovitskiy2016generating,gatys2016image,johnson2016perceptual} and a
discriminator feature matching loss~\cite{wang2018pix2pixHD}. Please see the
supplemental material for details.

% We use $\LearnedDownsamplingFactor=16$ and adapt \DownsamplingFactor such that the
% smaller dimension of $x_l$ is of size 256, so \TotalDownsamplingFactor depends
% on the image size (e.g. \TotalDownsamplingFactor $=64$ for image of size $1024
% \times 1024$).

% ----------- old stuff below

\section{Experiments}
\begin{figure*}[t]%[thp]
	\centering
	\captionsetup[subfigure]{labelformat=empty,justification=centering,aboveskip=1pt,belowskip=1pt}
	%%%%%%%%%%%%%%
	\begin{subfigure}[t]{0.24\textwidth}
		\centering
		\caption{$1024\times 1024$ label map}
		\includegraphics[width=0.98\linewidth]{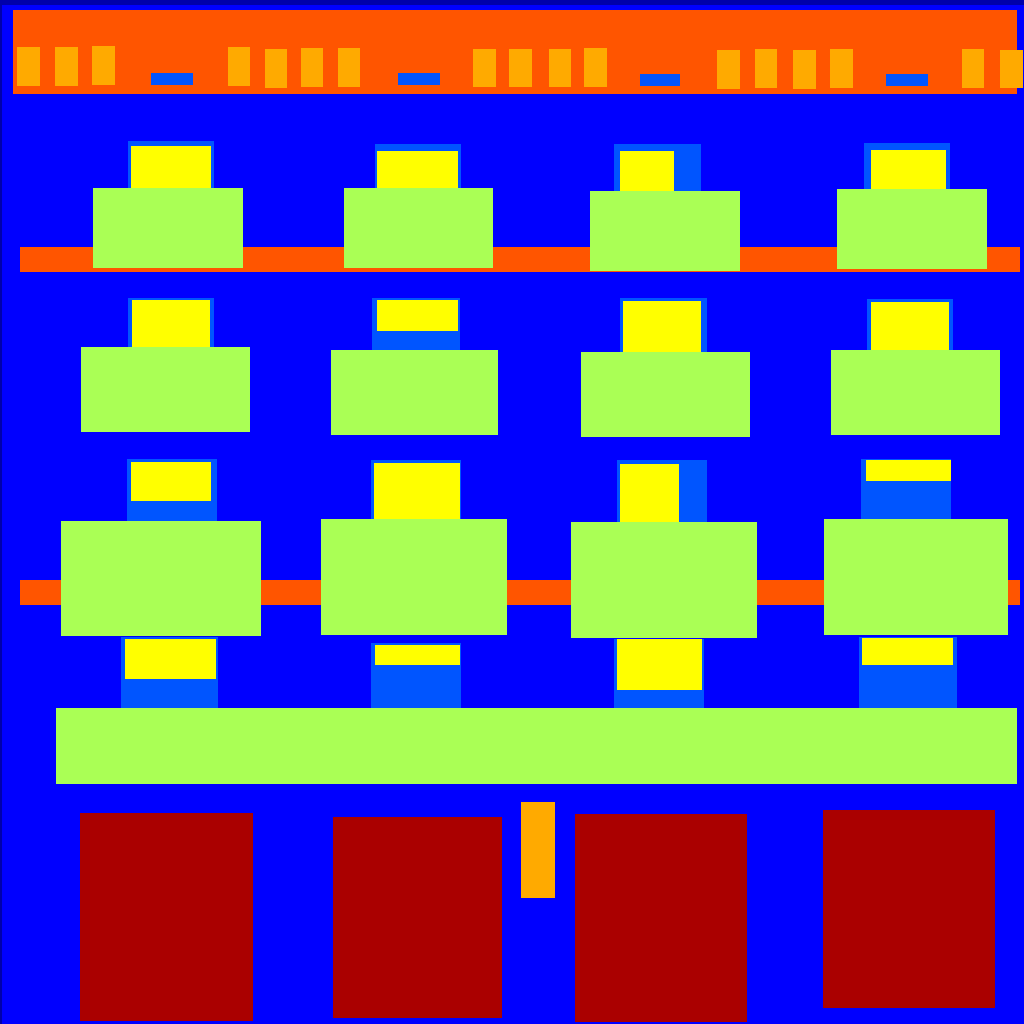}
	\end{subfigure}
	\begin{subfigure}[t]{0.24\textwidth}
		\centering
		\caption{SPADE~\cite{Park_2019_CVPR}, 359ms}
		\includegraphics[width=0.98\linewidth]{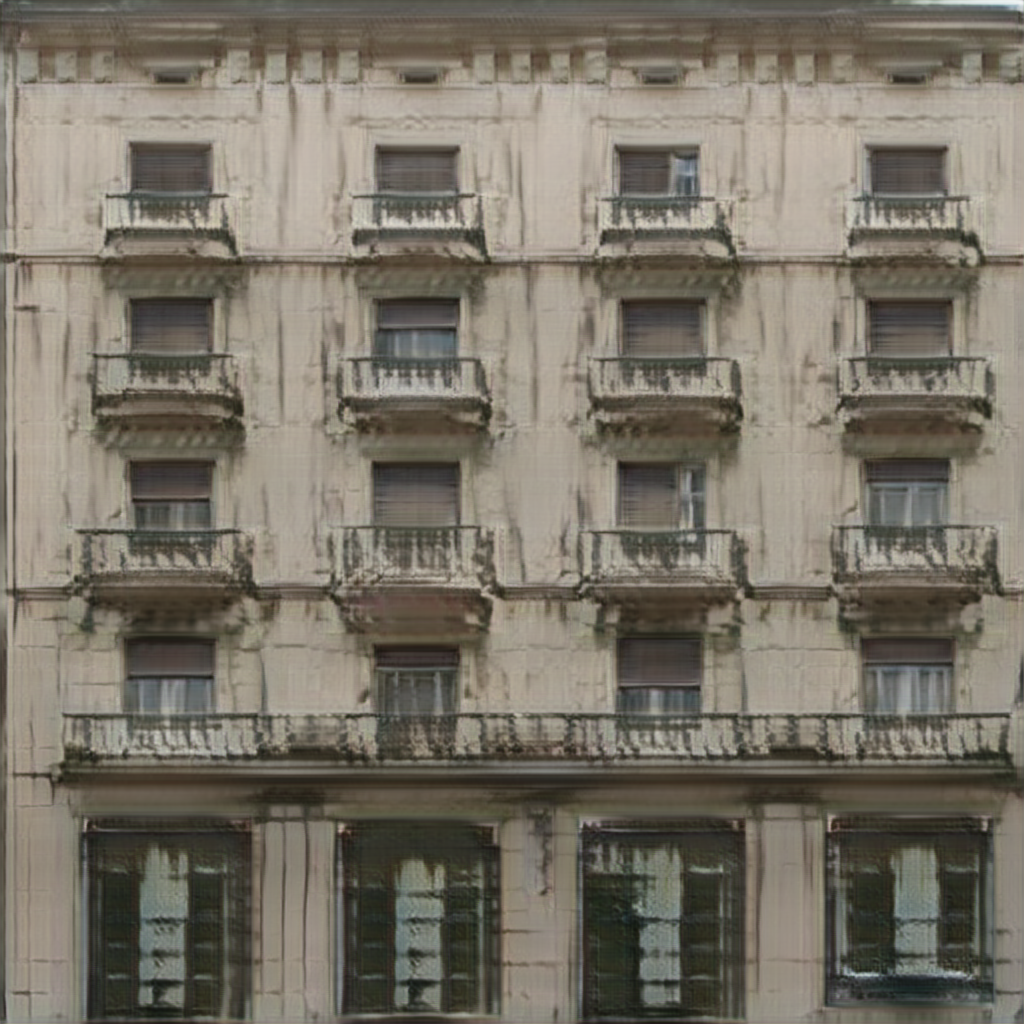}
	\end{subfigure}
	\begin{subfigure}[t]{0.24\textwidth}
		\centering
		\caption{pix2pixHD~\cite{wang2018pix2pixHD}, 166ms}
		\includegraphics[width=0.98\linewidth]{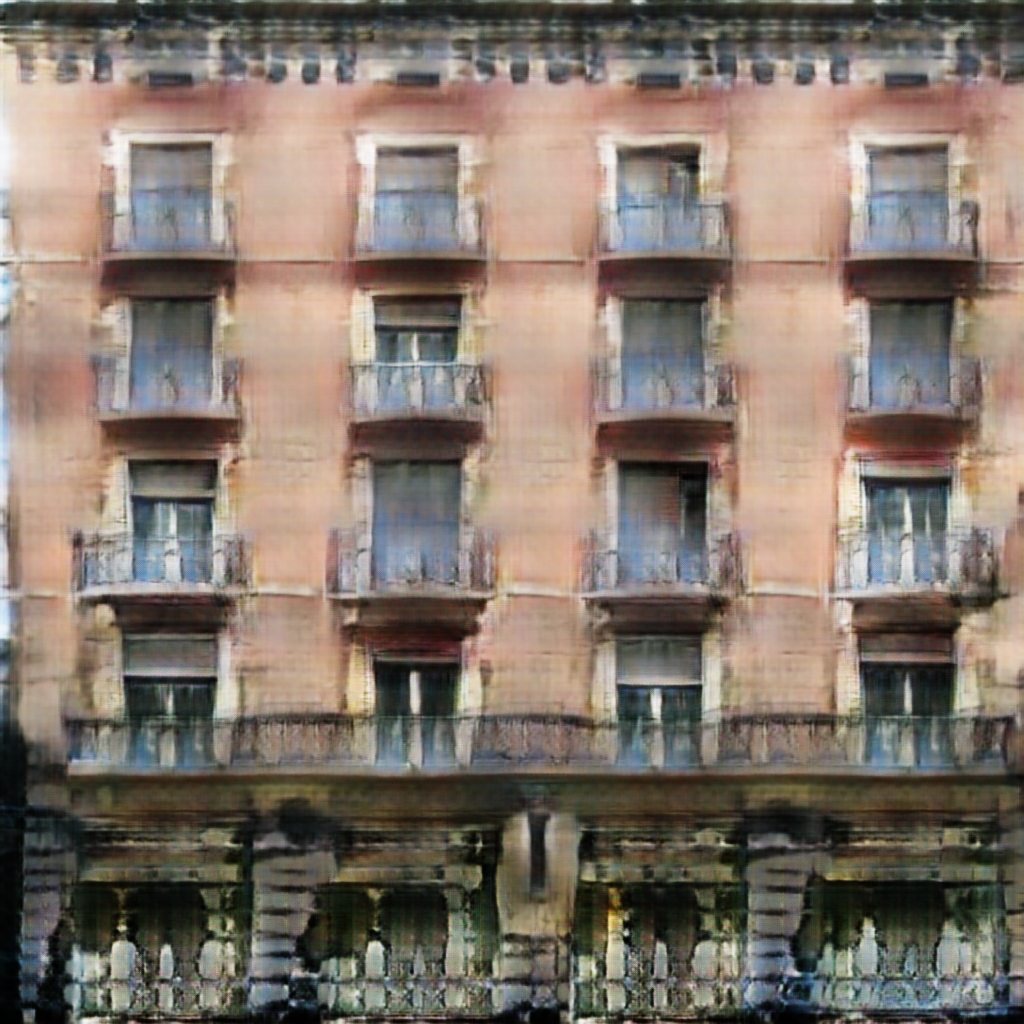}
	\end{subfigure}
	\begin{subfigure}[t]{0.24\textwidth}
		\centering
		\caption{ASAP (ours), 35ms}
		\includegraphics[width=0.98\linewidth]{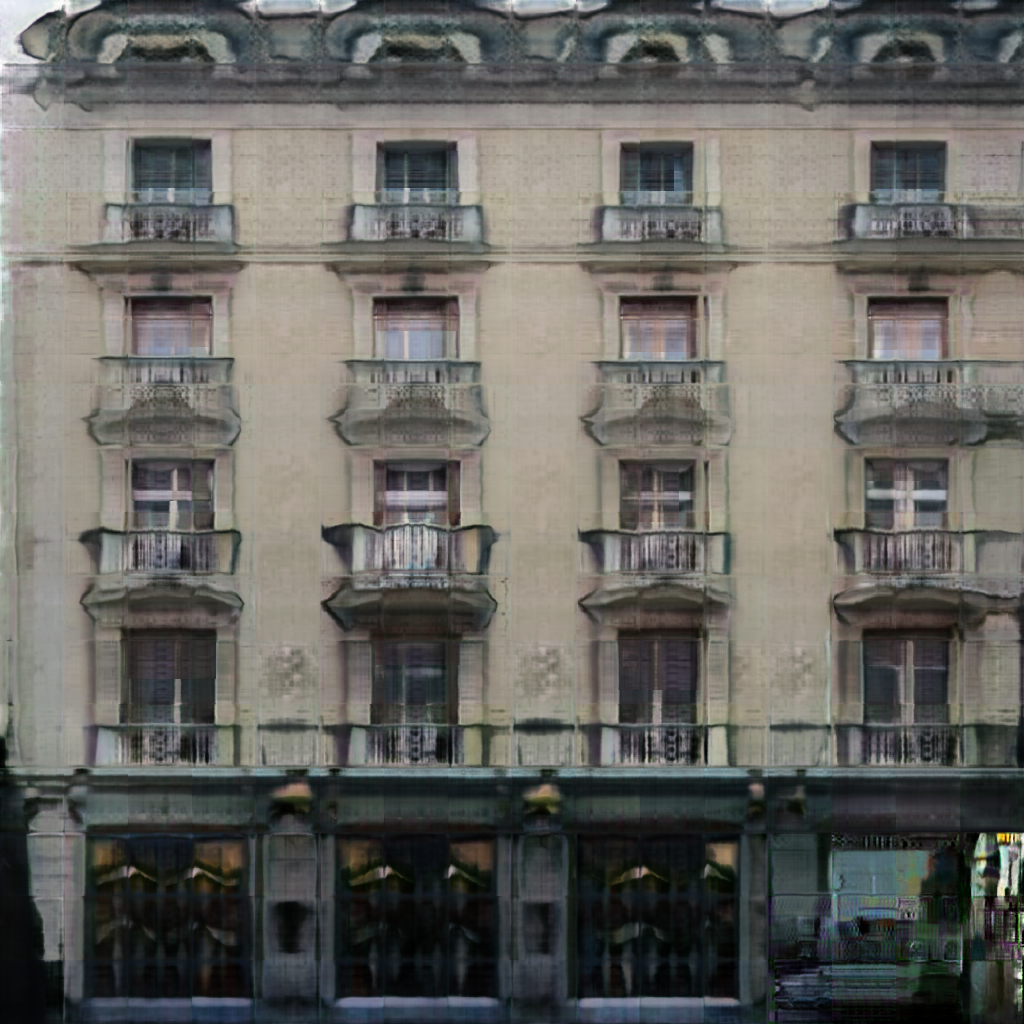}
	\end{subfigure}
	%%%%%%%%%%%%%%
	\begin{subfigure}[t]{0.24\textwidth}
		\centering
		%\caption{$1024\times 1024$ label map}
		\includegraphics[width=0.98\linewidth]{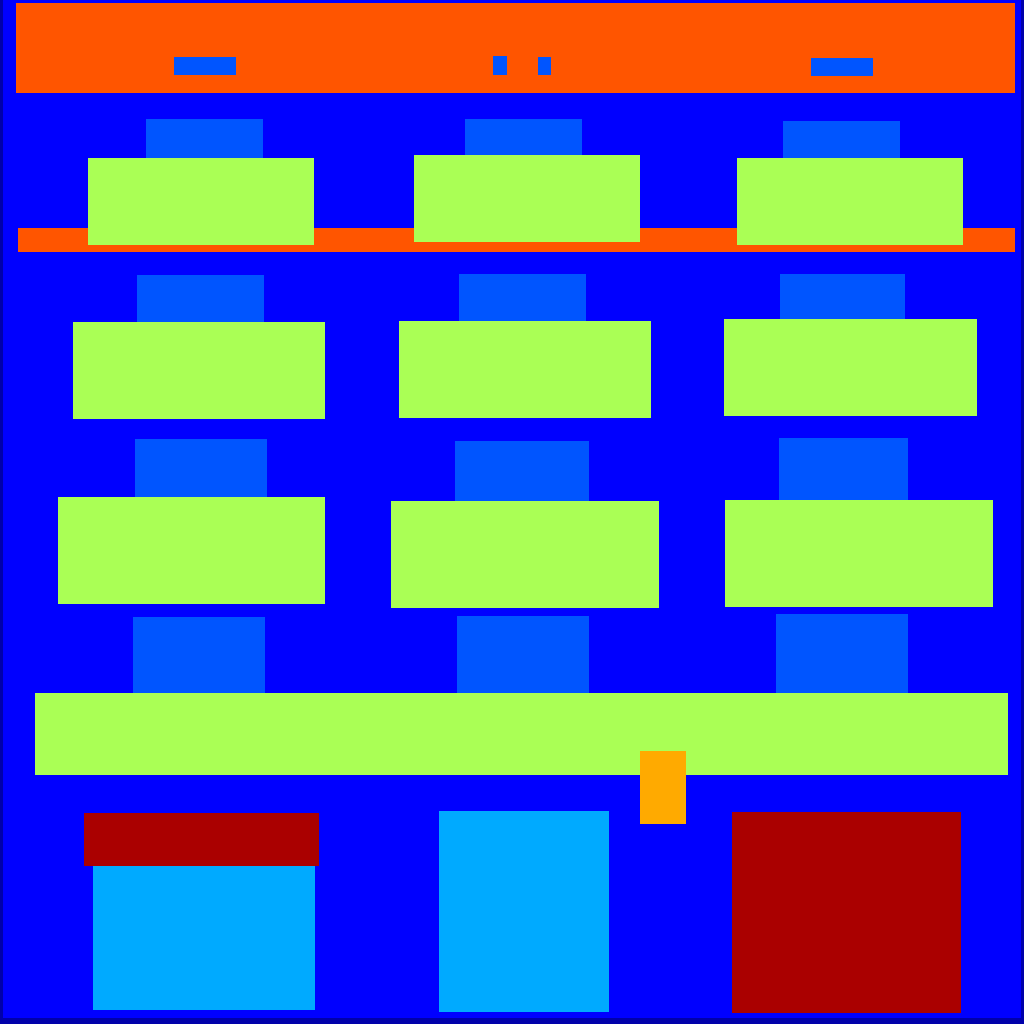}
	\end{subfigure}
	\begin{subfigure}[t]{0.24\textwidth}
		\centering
		%\caption{SPADE~\cite{Park_2019_CVPR}, 359ms}
		\includegraphics[width=0.98\linewidth]{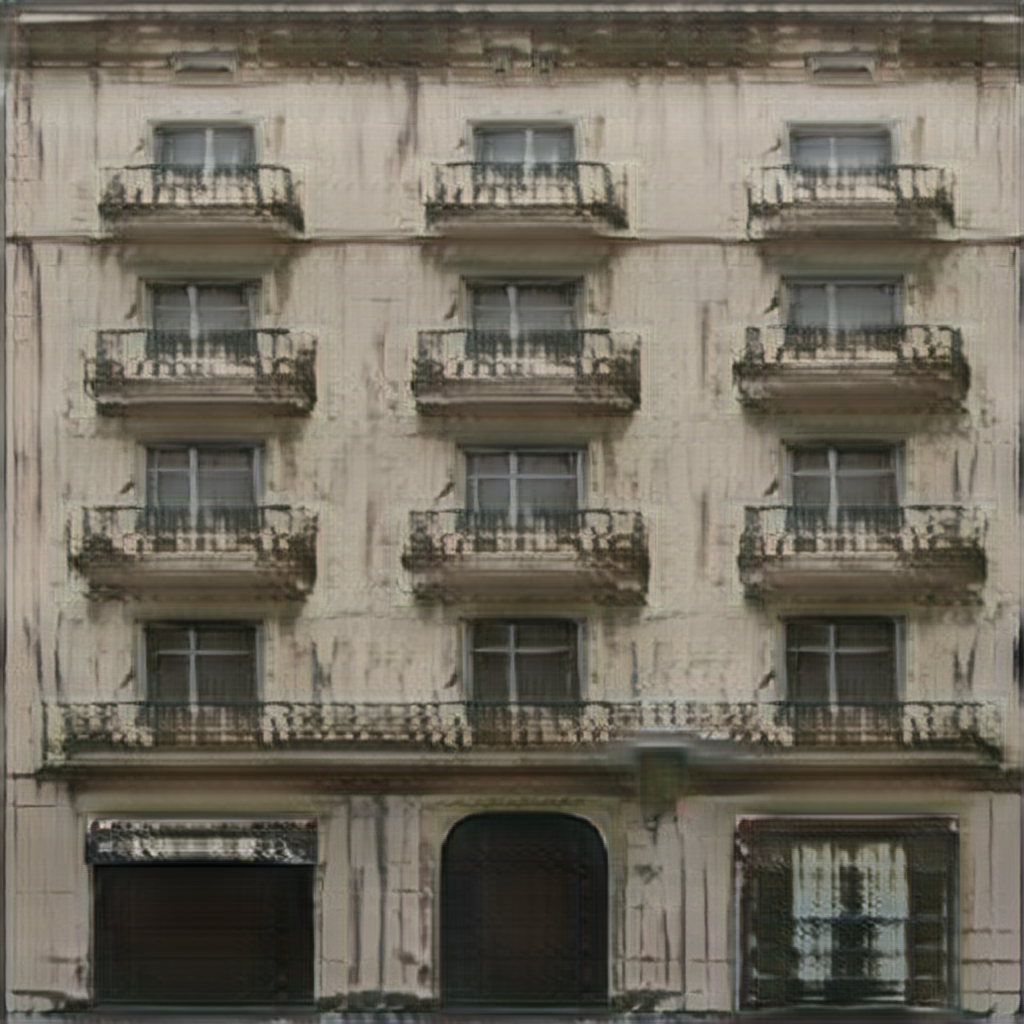}
		% \caption{vector sculpting}
	\end{subfigure}
	\begin{subfigure}[t]{0.24\textwidth}
		\centering
		%\caption{pix2pixHD~\cite{wang2018pix2pixHD}, 166ms}
		\includegraphics[width=0.98\linewidth]{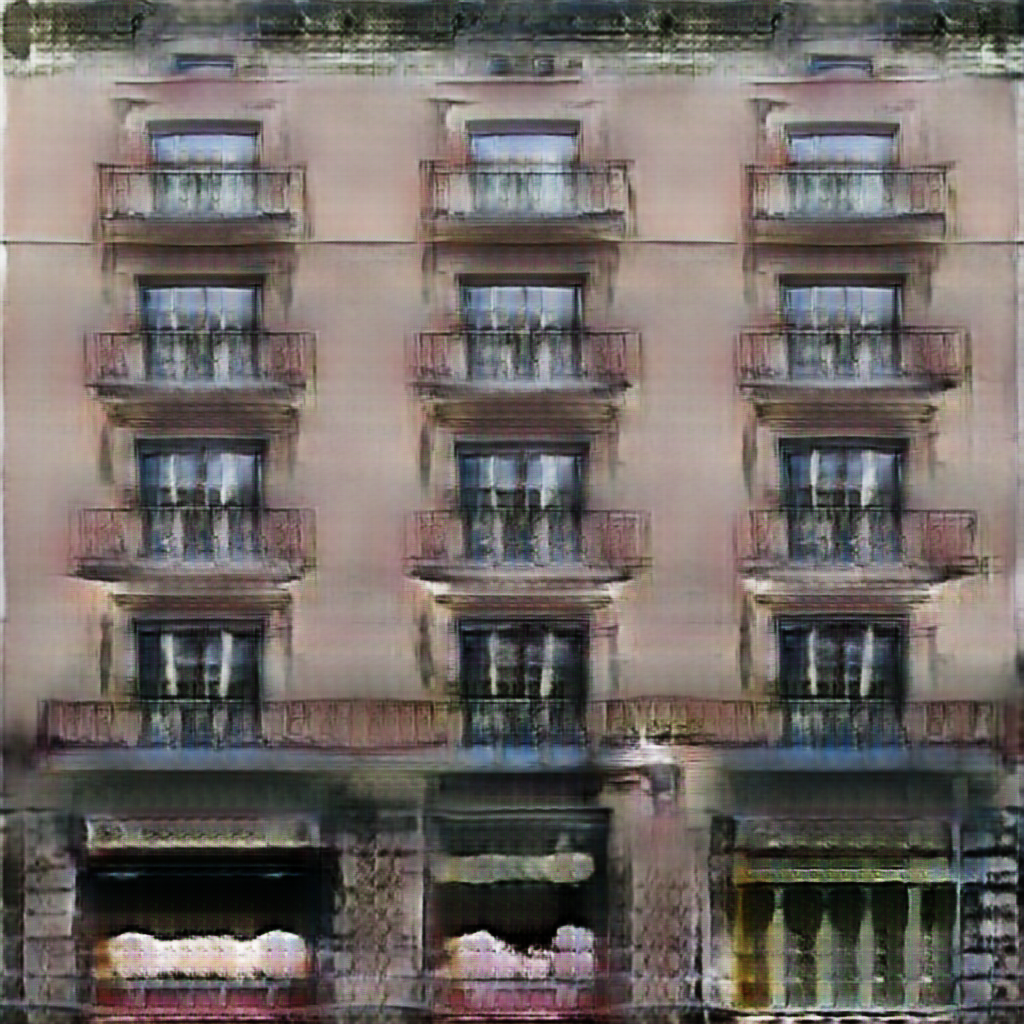}
		% \caption{vector sculpting}
	\end{subfigure}
	\begin{subfigure}[t]{0.24\textwidth}
		\centering
		%\caption{ASAP (ours), 35ms}
		\includegraphics[width=0.98\linewidth]{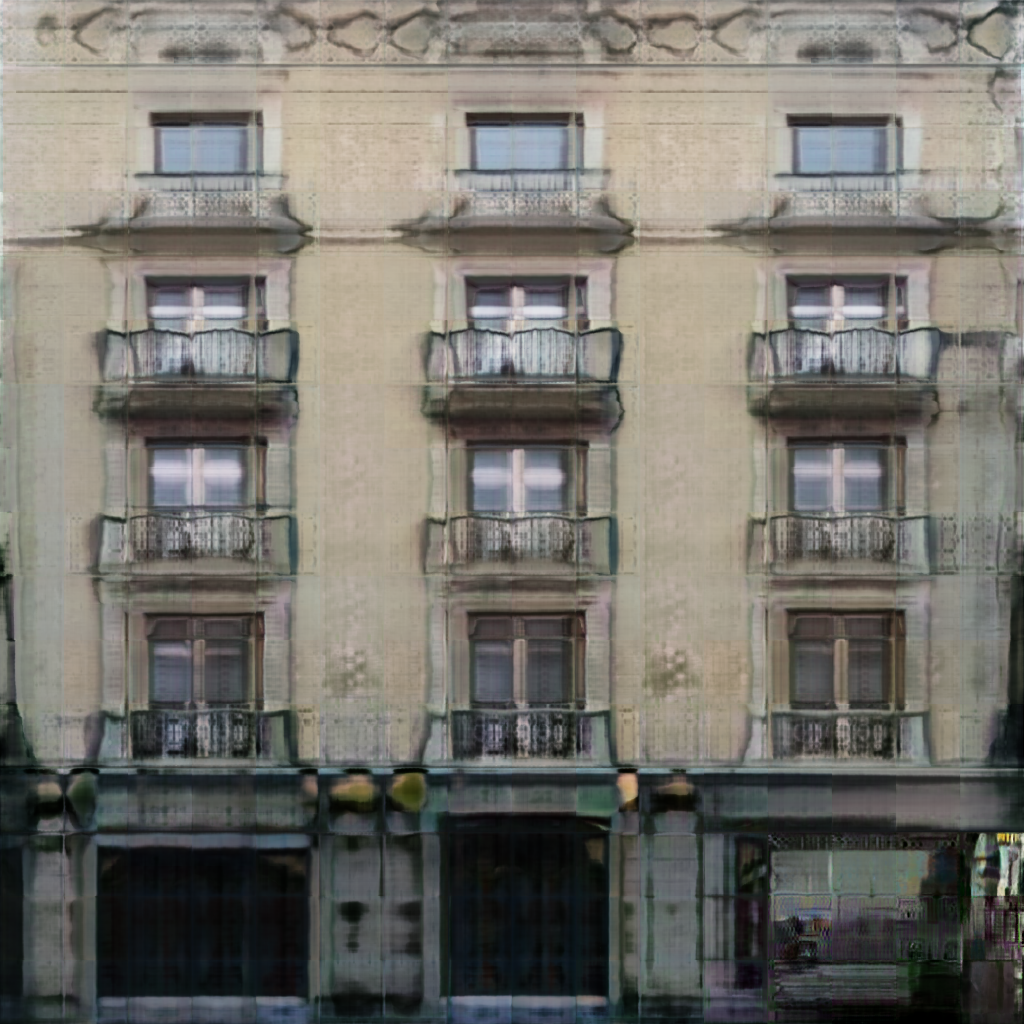}
		% \caption{vector sculpting}
	\end{subfigure}
	%%%%%%%%%%%%%%
	\begin{subfigure}[t]{0.19\textwidth}
		\centering
		\caption{$512\times 1024$ label map}
		\includegraphics[width=1\linewidth]{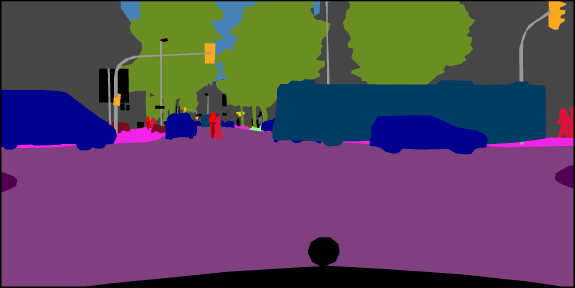}
	\end{subfigure}
	\begin{subfigure}[t]{0.19\textwidth}
		\centering
		\caption{CC-FPSE~\cite{liu2019learning}, \textbf{520ms}}
		\includegraphics[width=1\linewidth]{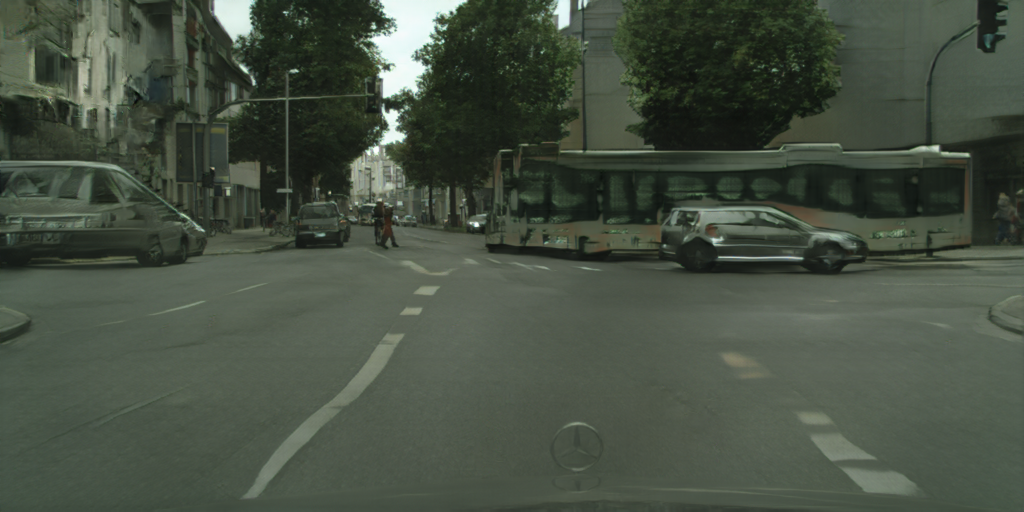}
		% \caption{vector sculpting}
	\end{subfigure}
	%\par\smallskip
	\begin{subfigure}[t]{0.19\textwidth}
		\centering
		\caption{SPADE~\cite{Park_2019_CVPR}, \textbf{190ms}}
		\includegraphics[width=1\linewidth]{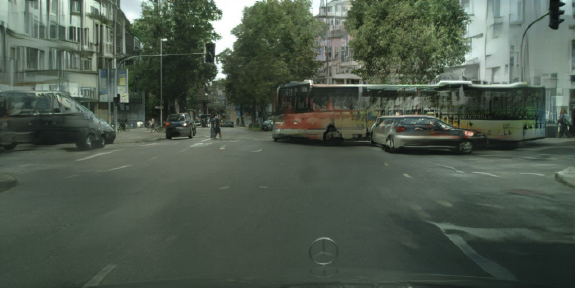}
	\end{subfigure}
	\begin{subfigure}[t]{0.19\textwidth}
		\centering
		\caption{pix2pixHD~\cite{wang2018pix2pixHD}, \textbf{95ms}}
		\includegraphics[width=1\linewidth]{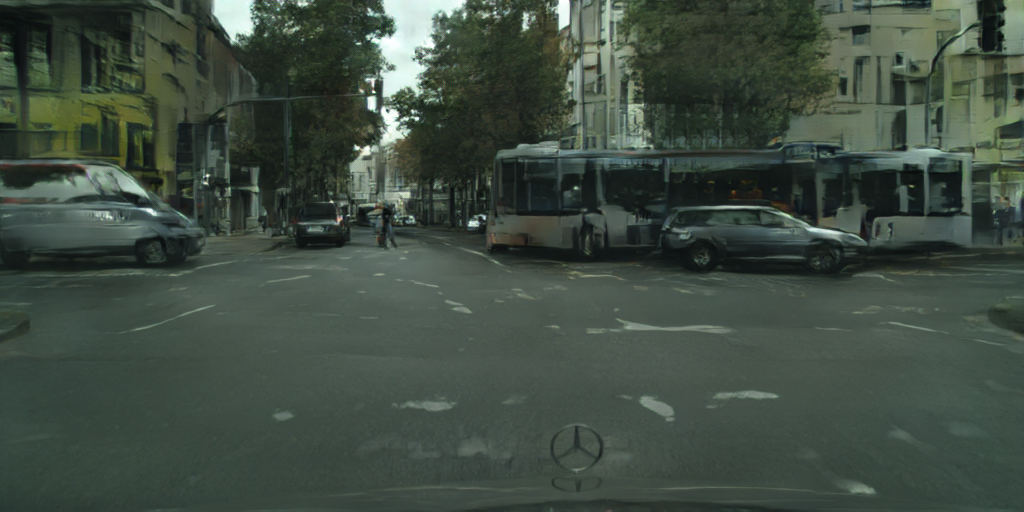}
	\end{subfigure}
	\begin{subfigure}[t]{0.19\textwidth}
		\centering
		\caption{ASAP (ours), \textbf{29ms}}
		\includegraphics[width=1\linewidth]{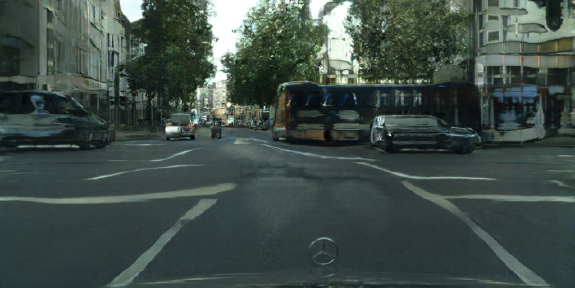}
	\end{subfigure}
	\par\smallskip
	\begin{subfigure}[t]{0.19\textwidth}
		\centering
		%\caption{$512\times 1024$ label map}
		\includegraphics[width=1\linewidth]{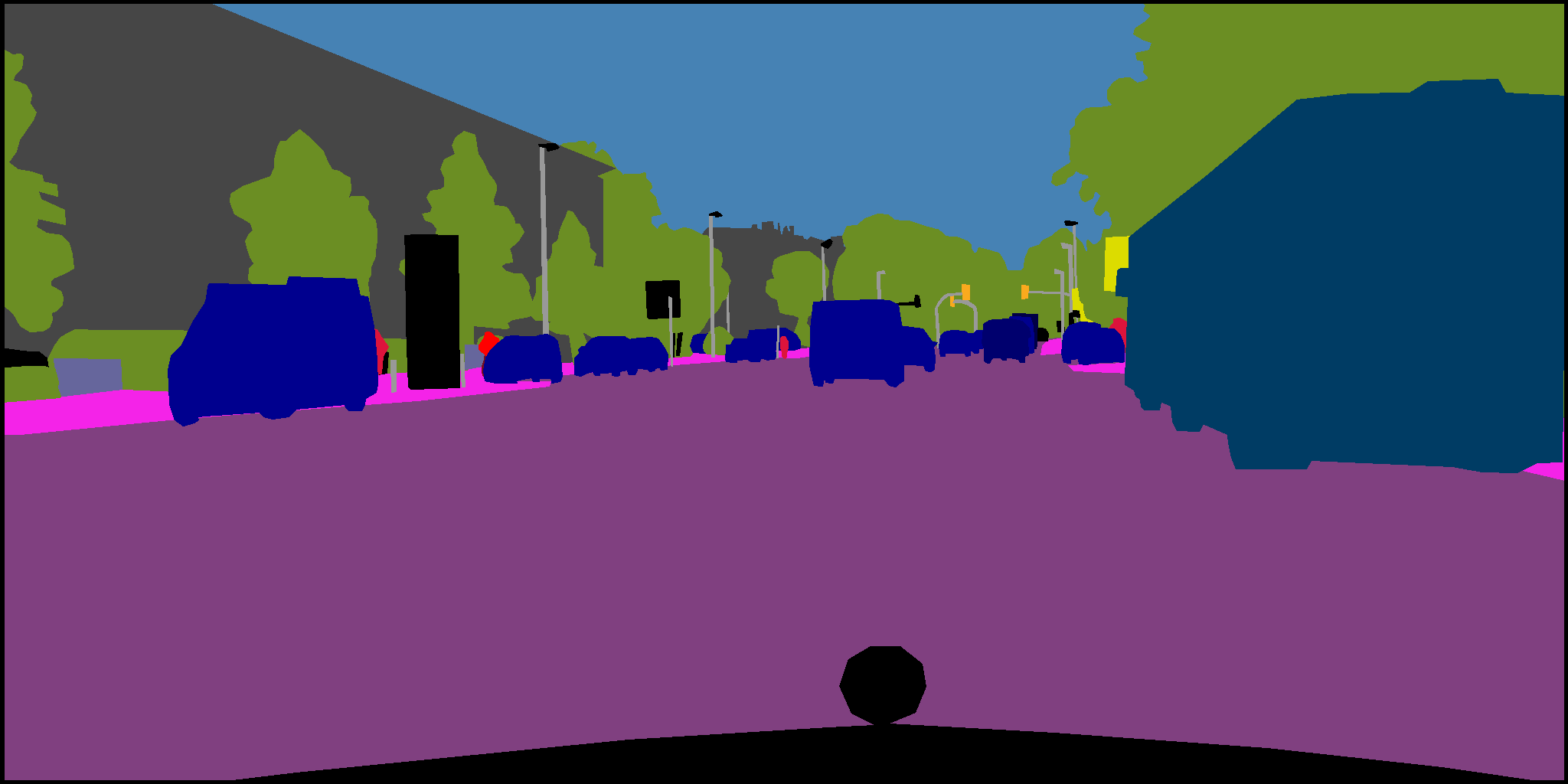}
	\end{subfigure}
	\begin{subfigure}[t]{0.19\textwidth}
		\centering
		%\caption{CC-FPSE~\cite{liu2019learning}, 512ms}
		\includegraphics[width=1\linewidth]{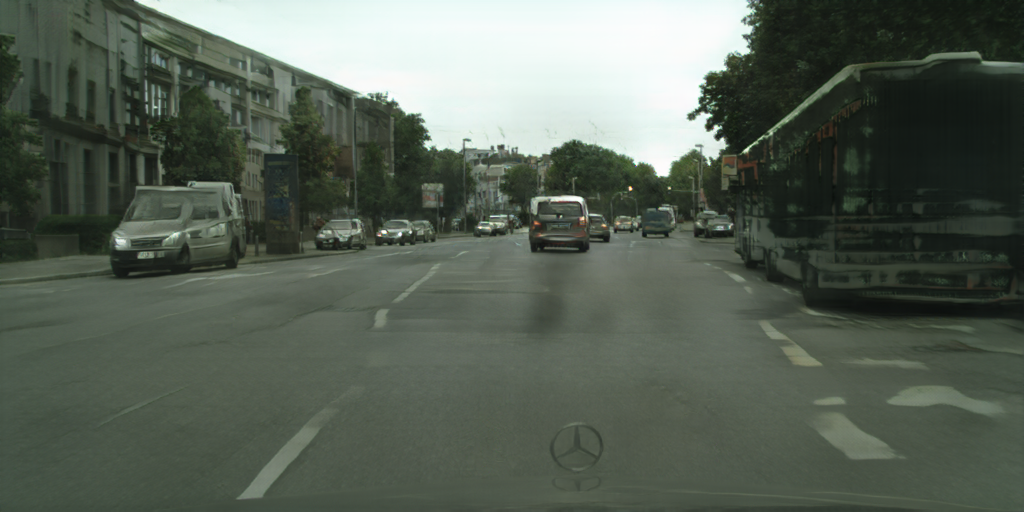}
		% \caption{vector sculpting}
	\end{subfigure}
	%\par\smallskip
	\begin{subfigure}[t]{0.19\textwidth}
		\centering
		%\caption{SPADE~\cite{Park_2019_CVPR}, 190ms}
		\includegraphics[width=1\linewidth]{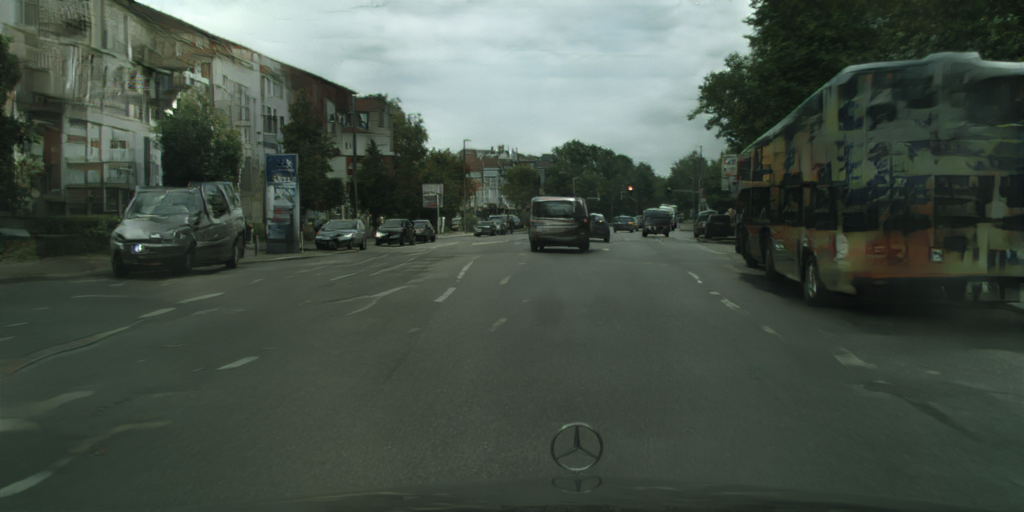}
	\end{subfigure}
	\begin{subfigure}[t]{0.19\textwidth}
		\centering
		%\caption{pix2pixHD~\cite{wang2018pix2pixHD}, 95ms}
		\includegraphics[width=1\linewidth]{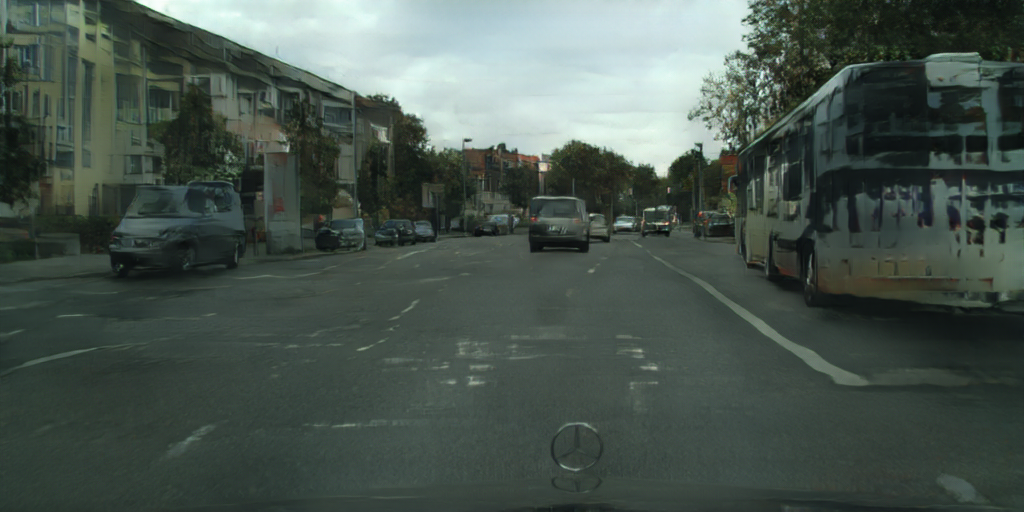}
	\end{subfigure}
	\begin{subfigure}[t]{0.19\textwidth}
		\centering
		%\caption{ASAP (ours), 29ms}
		\includegraphics[width=1\linewidth]{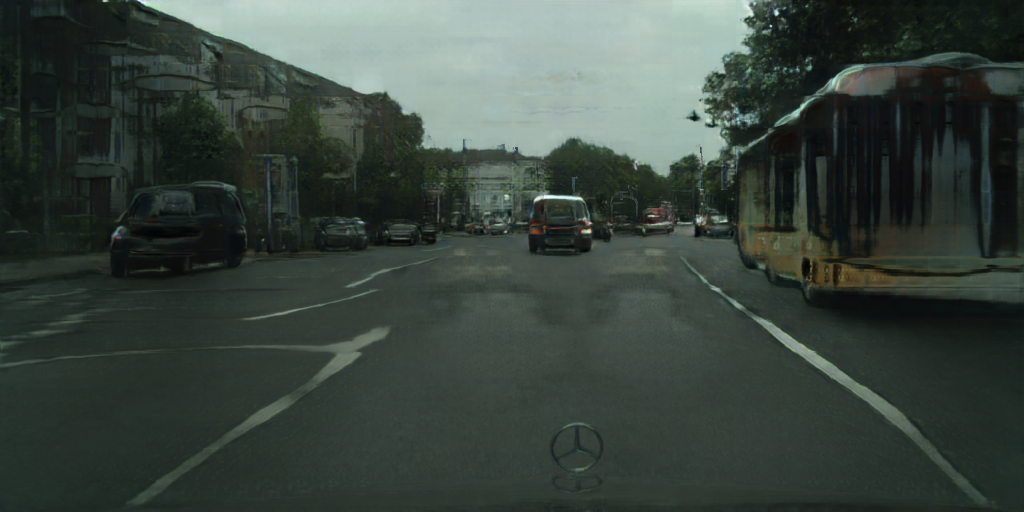}
	\end{subfigure}
	\par\smallskip
	\begin{subfigure}[t]{0.19\textwidth}
		\centering
		%\caption{$512\times 1024$ label map}
		\includegraphics[width=1\linewidth]{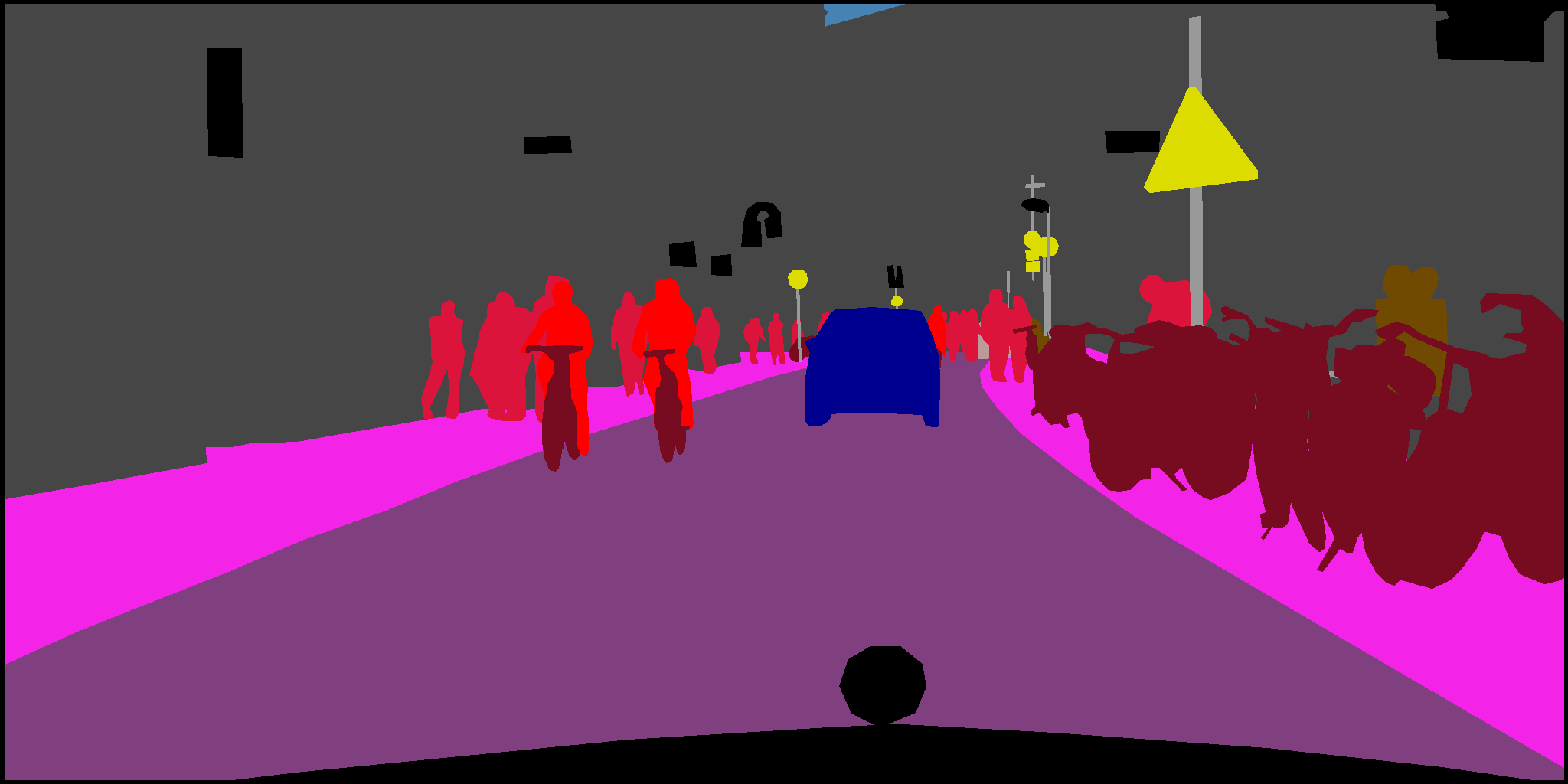}
	\end{subfigure}
	\begin{subfigure}[t]{0.19\textwidth}
		\centering
		%\caption{CC-FPSE~\cite{liu2019learning}, 512ms}
		\includegraphics[width=1\linewidth]{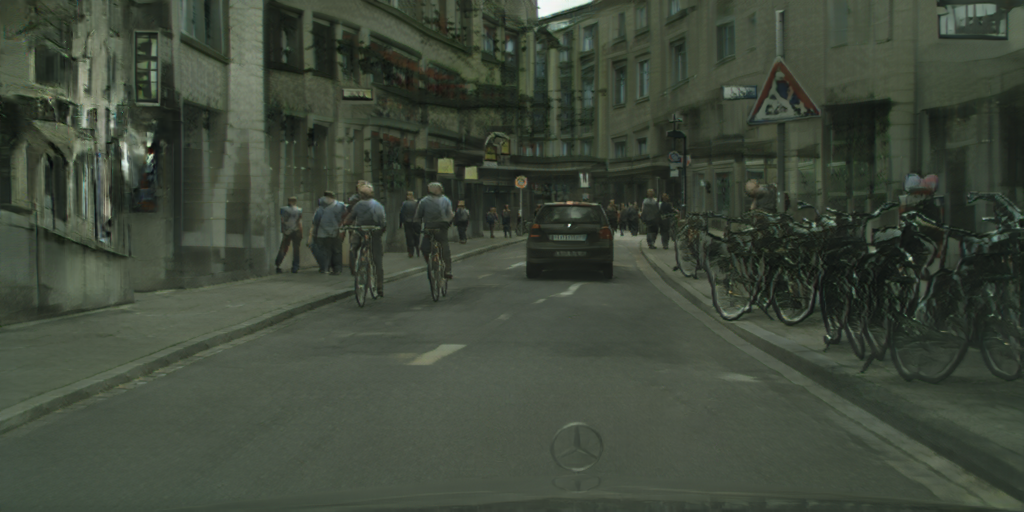}
		% \caption{vector sculpting}
	\end{subfigure}
	%\par\smallskip
	\begin{subfigure}[t]{0.19\textwidth}
		\centering
		%\caption{SPADE~\cite{Park_2019_CVPR}, 190ms}
		\includegraphics[width=1\linewidth]{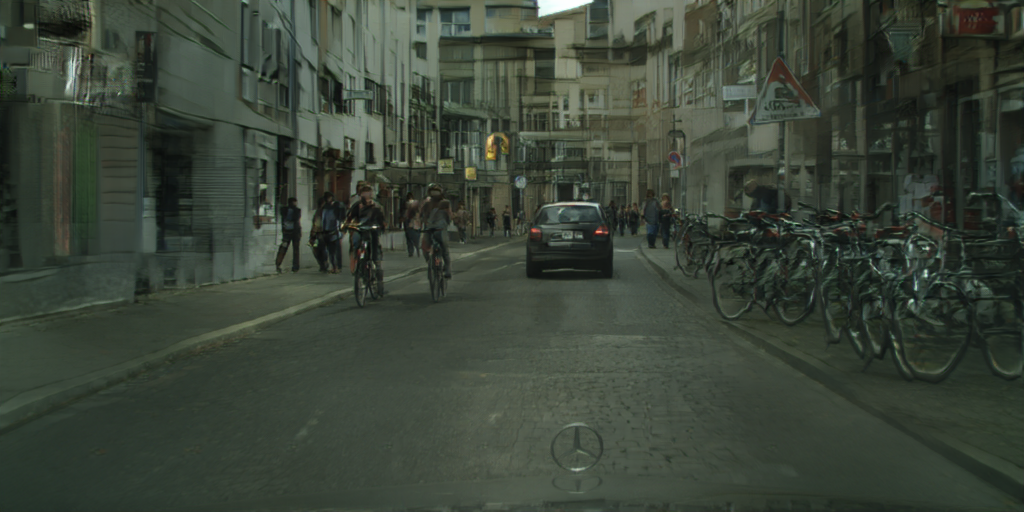}
	\end{subfigure}
	\begin{subfigure}[t]{0.19\textwidth}
		\centering
		%\caption{pix2pixHD~\cite{wang2018pix2pixHD}, 95ms}
		\includegraphics[width=1\linewidth]{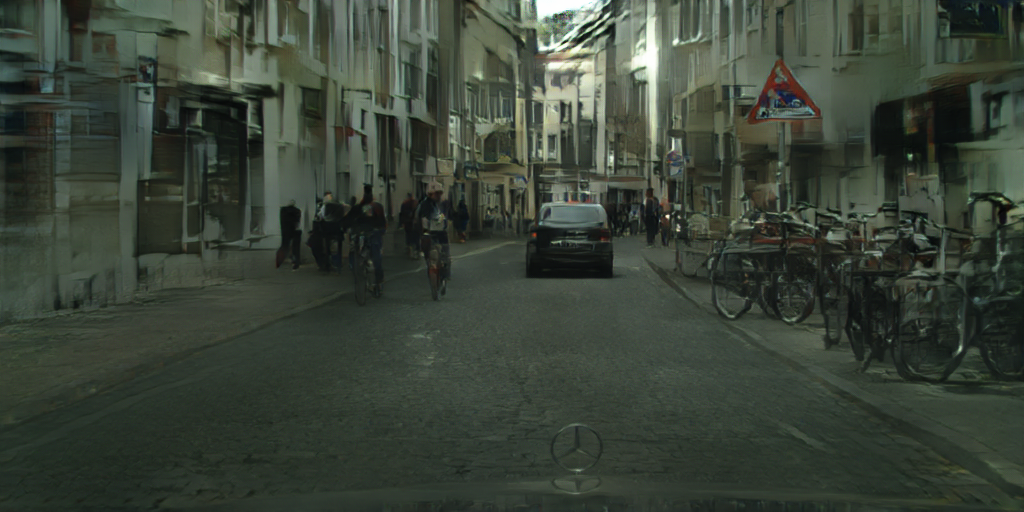}
	\end{subfigure}
	\begin{subfigure}[t]{0.19\textwidth}
		\centering
		%\caption{ASAP (ours), 29ms}
		\includegraphics[width=1\linewidth]{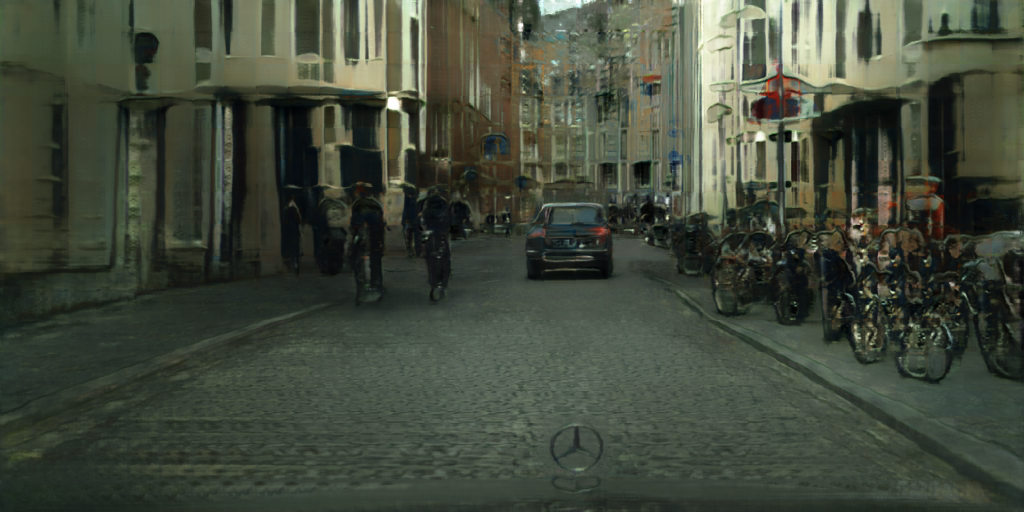}
	\end{subfigure}
	\caption{\label{fig:qualitative}\textbf{Qualitative comparison.} Our method (rightmost column) achieves similar visual quality as the baselines methods, for a fraction of the computational cost.
	}
\end{figure*}{}

We validate the performance of our method on several image translation tasks and compare runtime and image quality (quantitatively and
qualitatively) with previous work.

\myparagraph{Baselines.} 
We compare our model, \emph{ASAP-Net}, with five recent, competitive methods. 
These include three recent GAN based methods: 
\begin{inparaenum}[(i)]
\item CC-FPSE~\cite{liu2019learning}, which introduces new conditional convolution blocks that are aware of the different semantic regions, \item SPADE~\cite{Park_2019_CVPR}, which uses spatially adaptive normalization to translate semantic label maps into realistic images, and 
\item pix2pixHD~\cite{wang2018pix2pixHD}, which takes a residual learning approach that enables high-resolution image translation. We also include two earlier non-adverserial methods: 
\item SIMS~\cite{qi2018semi} which refines compositions of segments from a bank of training examples, and 
\item CRN~\cite{chen2017photographic} which uses a coarse-to-fine generation strategy.
\end{inparaenum}
Finally, we also include a comparison with the lighter-weight pix2pix
model~\cite{isola2017image}, which was the first to address this task.

\myparagraph{Datasets.}
We use three datasets:
\begin{inparaenum}[(i)] 
\item CMP Facades~\cite{tylevcek2013facades}, which contains 400 pairs of architecture labels and their corresponding building images of varying sizes. We randomly split it into 360/40 training/validation images, respectively.
We construct two versions of this dataset, corresponding to resolutions  $512 \times 512$ and $1024\times1024$. 
\item Cityscapes~\cite{cordts2016cityscapes}, which contains images of urban scenes and their semantic label maps, split into 3000/500
training/validation images. We construct $256\times512$ and $512\times1024$ versions of this dataset. 
\item NYU depth dataset~\cite{silberman2012indoor}, which contains 1449 pairs of aligned RGB and depth images of indoor scenes, split into 1200/249 training/validation images.
\end{inparaenum}

\subsection{Evaluations}
\begin{figure*}[t]
    \centering
    %\begin{subfigure}[]{0.43\textwidth}    \includegraphics[width=0.99\hsize]{figs/2afc/2afc_vert_v6.pdf}
    %\caption{User preference for our images }\label{fig:eval:amt}
    %\end{subfigure}
    %\hspace{0.05\textwidth}
    %\begin{subfigure}[]{0.51\textwidth}
    %\includegraphics[width=1\hsize]{figs/acc/acc_vs_runtime_v5_log.pdf}
    %\caption{\label{fig:eval:acc}Pixel accuracy vs.~run-time}
    %\end{subfigure}
    %\par\smallskip
    %\begin{subfigure}[]{1\textwidth}
    \includegraphics[width=1\hsize]{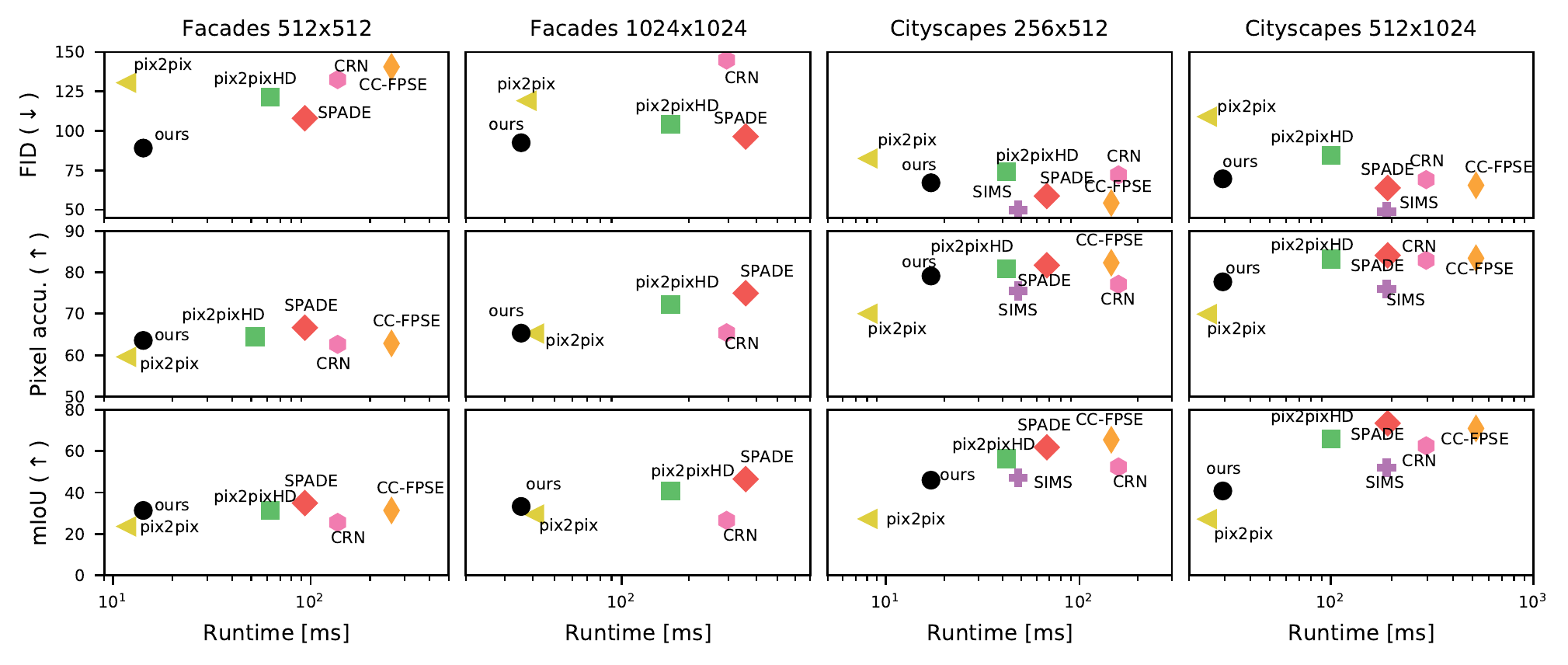}
    %\end{subfigure}
    \caption{\textbf{Performance vs.~runtime}
    %\subref{fig:eval:fid},\subref{fig:eval:acc}
    We compare our method to all baselines using several metrics. FID (lower is better), segmentation accuracy and mean intersection over union (higher is better) show our model outperforms pix2pix and is comparable to CRN, SIMS, pix2pixHD, SPADE, CC-FPSE across all datasets and resolutions.
    This is while our model is nearly as fast as pix2pix and significantly faster than
    all other baselines.
    %\tamar{[not sure about log scale. it makes pix2pix look much faster than ours (it is only ~5ms faster)]}
    %
} \label{fig:eval} 

\end{figure*}

\begin{figure}[htp]
    \centering  \includegraphics[width=1\columnwidth]{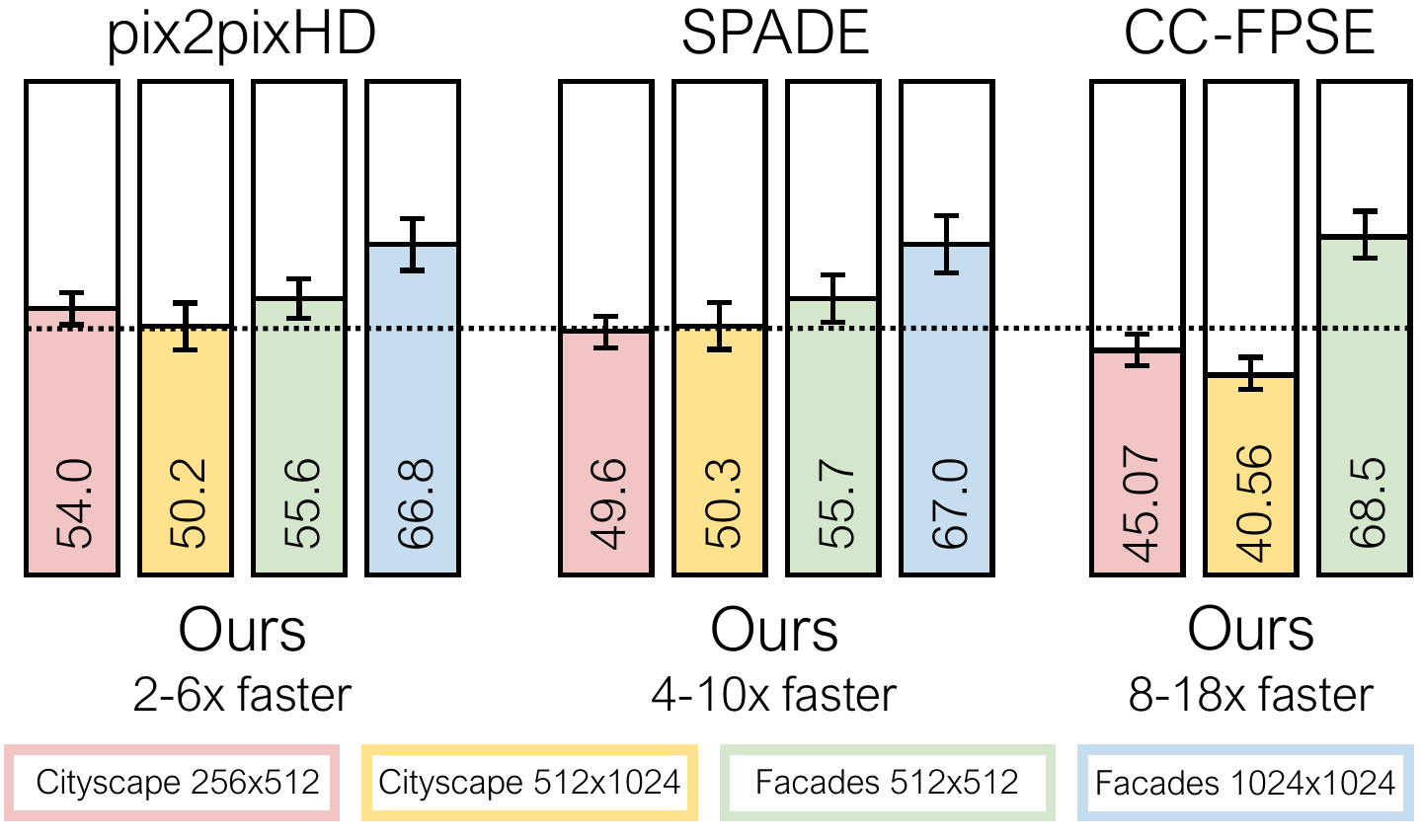}
    %\caption{User preference for our images }\label{fig:eval:amt}
    \vspace{-0.5cm}
    \caption{\textbf{Human Perceptual Study.}
     We ask users to choose the most realistic result in a Two
    Alternative Forced Choice (2AFC) perceptual test.
    We compare our outputs independently to pix2pixHD, SPADE and CC-FPSE.
    Across most datasets and resolutions, users prefer our images with a rate over 50\%.
    On the high-resolution Facades dataset user preference for our method
    even reaches 67\%.}
    \vspace{-0.3cm}
    \label{fig:amt}
\end{figure}

%We first analyze the runtime of our algorithm and observe large speedups relative to baselines. Our method maintains similar visual quality, as assessed by human perceptual judgments. We further dissect the results with automatic, quantitative measures.

We compare our method to the baselines in terms of runtime, human perceptual judgements, and automatic metrics.

\begin{figure}[htp]
    \centering
    \begin{subfigure}[]{1\columnwidth}
    \centering
    \includegraphics[width=1\hsize]{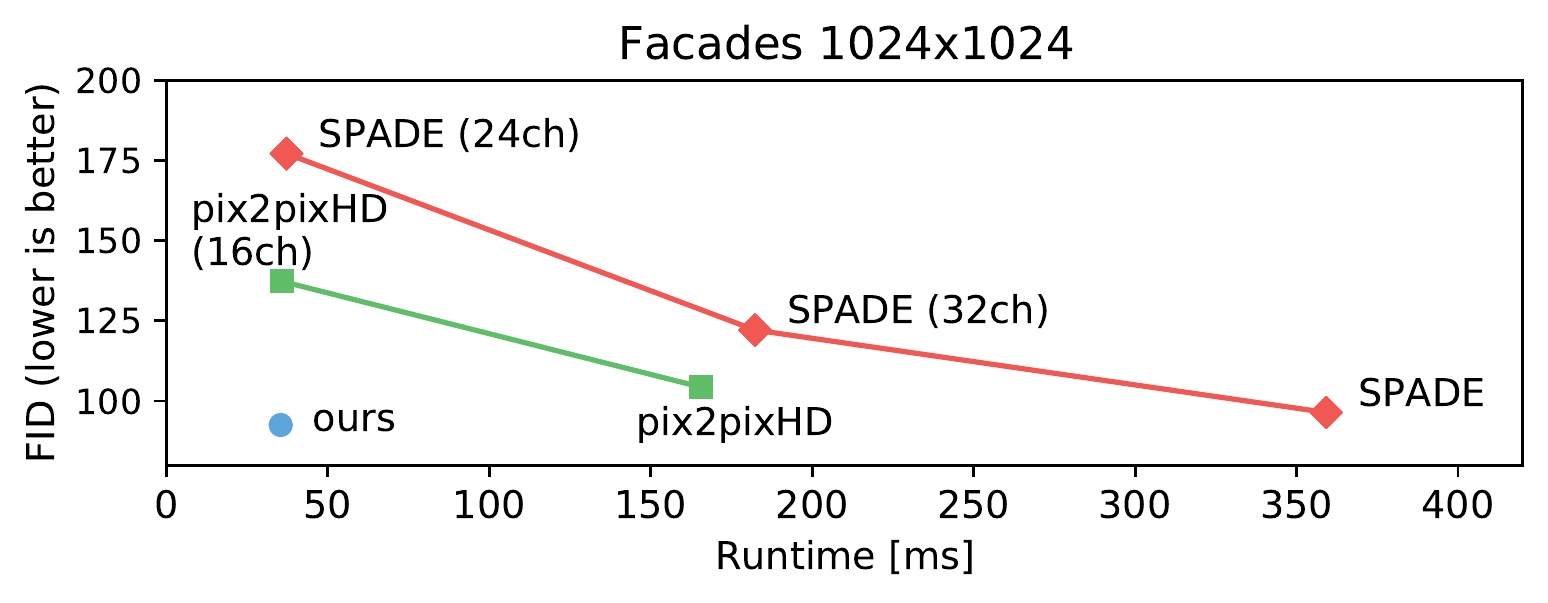}
    \end{subfigure}
    \caption{\label{fig:fid_curve}\textbf{ASAP vs.~slimmer baseline variants.} When gradually decreasing the number of channels in SPADE~\cite{Park_2019_CVPR} and pix2ixHD~\cite{wang2018pix2pixHD} until reaching the execution time of our ASAP network, both experience a severe degradation in FID.   
    %\todo{add slim p2phd}
    %When gradually cutting down the number of parameters of the SPADE model~\cite{Park_2019_CVPR} to have the same runtime as our ASAP model, SPADE incurs degradation in FID scores. for both Facades $512 \times 512$ images, please see SM for additional comparisons. \label{fig:fid_vs_runtime}
    }
\end{figure}{}
\begin{figure*}[htp]
	\centering
	\captionsetup[subfigure]{justification=centering,aboveskip=1pt,belowskip=1pt}
	\begin{subfigure}[t]{0.16\textwidth}
		\centering
		\caption{\label{fig:abl:input}label map}
		\includegraphics[width=0.98\linewidth]{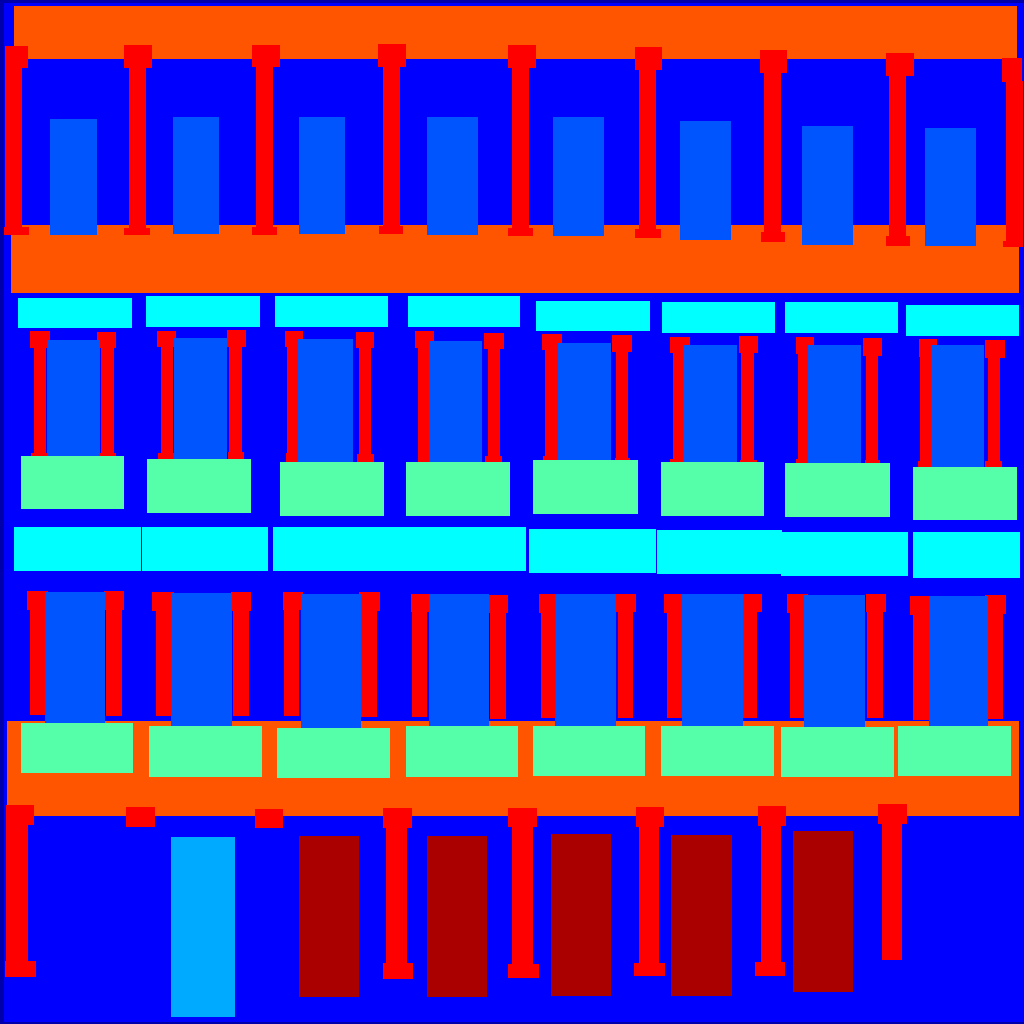}
	\end{subfigure}
	\begin{subfigure}[t]{0.16\textwidth}
		\centering
		\caption{\label{fig:abl:pos_encoding}w/o pos. enc.}
		\includegraphics[width=0.98\linewidth]{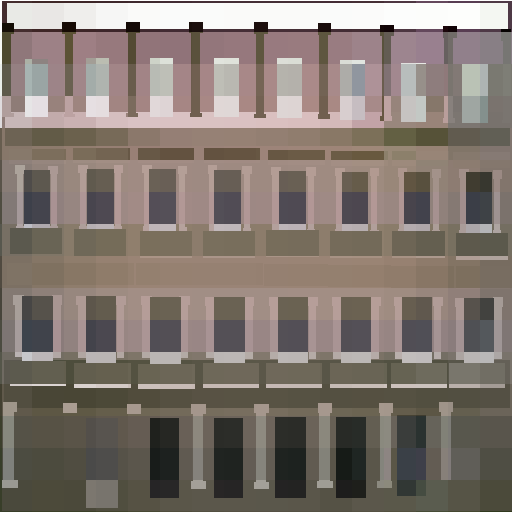}
	\end{subfigure}
	\begin{subfigure}[t]{0.16\textwidth}
		\centering
		\caption{\label{fig:abl:constant}constant $\HighresNet_\Pixel$}
		\includegraphics[width=0.98\linewidth]{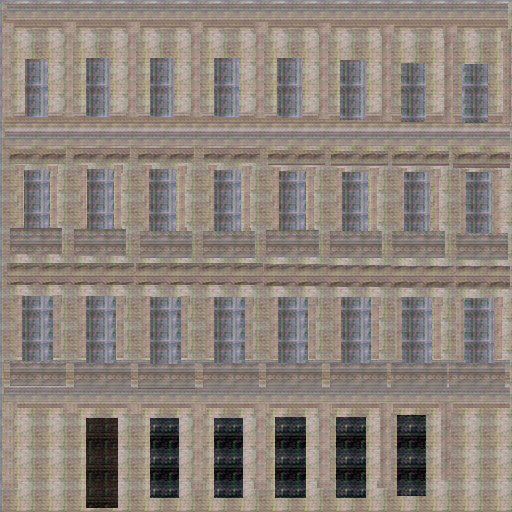}
	\end{subfigure}
	\begin{subfigure}[t]{0.16\textwidth}
		\centering
		\caption{\label{fig:abl:speed1}\TotalDownsamplingFactor $=64$}
		\includegraphics[width=0.98\linewidth]{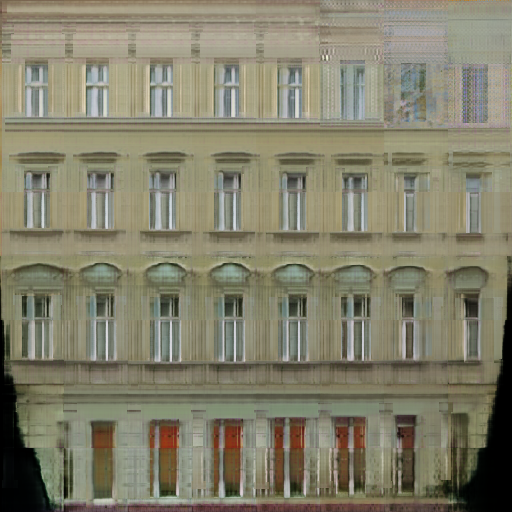}
	\end{subfigure}
	\begin{subfigure}[t]{0.16\textwidth}
		\centering
		\caption{\label{fig:abl:speed2}\TotalDownsamplingFactor $=16$}
		\includegraphics[width=0.98\linewidth]{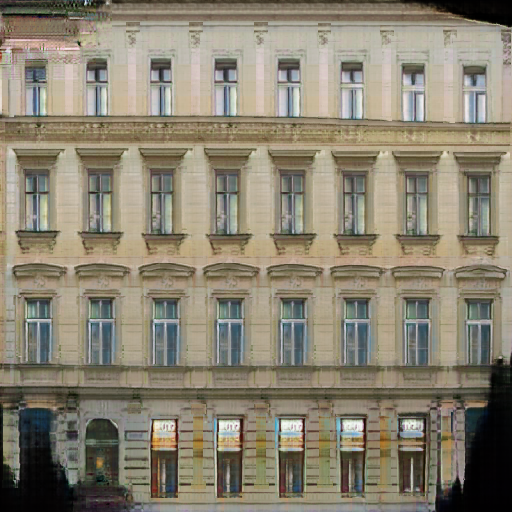}
	\end{subfigure}
	\begin{subfigure}[t]{0.16\textwidth}
		\centering
		\caption{\label{fig:abl:ours}ours, \TotalDownsamplingFactor$=32$}
		\includegraphics[width=0.98\linewidth]{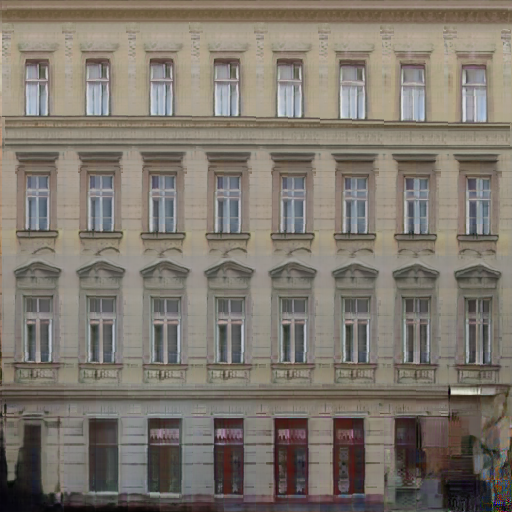}
	\end{subfigure}
	\caption{\label{fig:ablation}\textbf{Ablations.} The two main
		components of our model are crucial for its success.
		\subref{fig:abl:pos_encoding} Omitting the
		position encoding, $\HighresNet_\Pixel$ can  only use the semantic
		labels pointwise and is unable to hallucinate new high frequency details.
		\subref{fig:abl:constant}~When preventing
		$\HighresNet_\Pixel$ from varying spatially, the
		expressiveness of the model is severely limited.
		\subref{fig:abl:speed1}-\subref{fig:abl:speed2} The down-sampling
		factor $\TotalDownsamplingFactor$ is critical for both quality and speedup;
		using larger $\TotalDownsamplingFactor$ reduces the runtime of the model
		($12.1$ms) but is detrimental to visual quality, whereas reducing
		$\TotalDownsamplingFactor$ results in high quality image, but with
		$2\times$ longer runtime ($28.1$ms). \subref{fig:abl:ours} Our final
		model strikes a good balance. It can translate the $512 \times 512$
		inputs \subref{fig:abl:input} into realistic images in only~$14.2$ms. 
	}
\end{figure*}{}

\begin{figure}[htp]
	\centering
	\captionsetup[subfigure]{labelformat=empty,justification=centering,aboveskip=1pt,belowskip=1pt}
	\begin{subfigure}[t]{0.24\columnwidth}
		\centering
		\caption{input image}
		\includegraphics[width=0.95\linewidth]{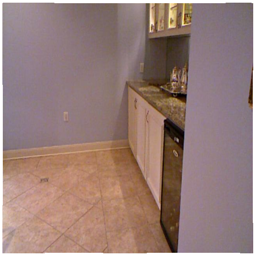}
	\end{subfigure}\begin{subfigure}[t]{0.24\columnwidth}
		\centering
		\caption{SPADE}
		\includegraphics[width=0.95\linewidth]{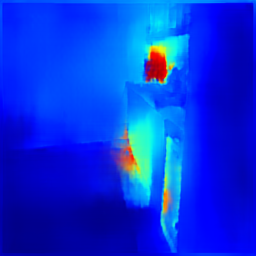}
	\end{subfigure}
	\begin{subfigure}[t]{0.24\columnwidth}
		\centering
		\caption{ASAP-Net}
		\includegraphics[width=0.95\linewidth]{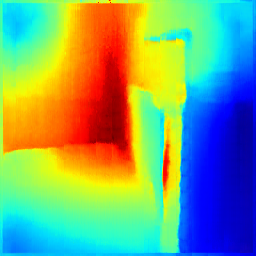}
	\end{subfigure}
	\begin{subfigure}[t]{0.24\columnwidth}
		\centering
		\caption{ground truth}
		\includegraphics[width=0.95\linewidth]{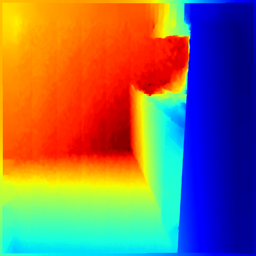}
	\end{subfigure}
	\\
	\begin{subfigure}[t]{0.24\columnwidth}
		\centering
		\includegraphics[width=0.95\linewidth]{figs/depth_images/01261_in.png}
	\end{subfigure}\begin{subfigure}[t]{0.24\columnwidth}
		\centering
		\includegraphics[width=0.95\linewidth]{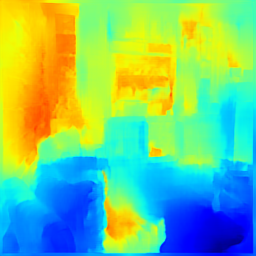}
	\end{subfigure}
	\begin{subfigure}[t]{0.24\columnwidth}
		\centering
		\includegraphics[width=0.95\linewidth]{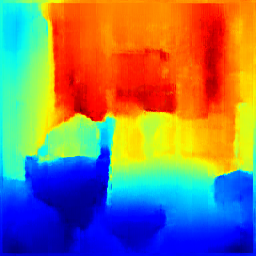}
	\end{subfigure}
	\begin{subfigure}[t]{0.24\columnwidth}
		\centering
		\includegraphics[width=0.95\linewidth]{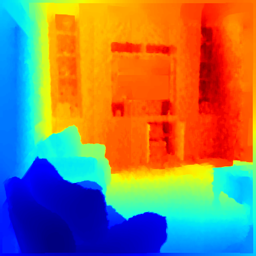}
	\end{subfigure}
	%\begin{subfigure}[t]{0.32\columnwidth}
	%	\centering
	%	\caption{input image}
	%	\includegraphics[width=0.95\linewidth]{figs/depth_images/01348_in.png}
	%\end{subfigure}
	%\begin{subfigure}[t]{0.32\columnwidth}
	%	\centering
	%	\caption{ASAP-Net}
	%	\includegraphics[width=0.95\linewidth]{figs/depth_images/01348_ASAP.png}
	%\end{subfigure}
	%\begin{subfigure}[t]{0.32\columnwidth}
	%	\centering
	%	\caption{ground truth}
	%	\includegraphics[width=0.95\linewidth]{figs/depth_images/01348_GT.png}
	%\end{subfigure}
	\caption{\label{Fig:depth} \textbf{Beyond labels.} The architecture of SPADE \cite{Park_2019_CVPR} is designed specifically for the task of translating labels into images, and is therefore less successful for task like depth estimations. Our approach, on the other hand, manages to preform well on this task. Please see SM for more results. 
	%the task Example for test results of ASAP-Net trained on the NYU dataset~\cite{Silberman:ECCV12}. \todo{[add SPADE results.]}
	}
	\vspace{-0.5cm}
\end{figure}

\myparagraph{Speed.}\label{sec:speed}
We benchmark the inference time of all models on an Nvidia GeForce 2080ti GPU.
Working at low resolution with only pointwise full-resolution operations gives our model up to a $6\times$ speedup compared to pix2pixHD, $10\times$ speedup compared to SPADE (in spite of the comparable number of trainable parameters), SIMS~\cite{qi2018semi} and CRN~\cite{chen2017photographic}, and $18\times$ speedup compared to CC-FPSE~\cite{liu2019learning}, depending on resolution.
Figure~\ref{fig:fig_0} summarizes inference times for various input sizes and shows that our runtime advantage becomes more significant at higher resolutions.
One of the reasons for this is that the runtime of our low resolution convolutional stream is almost constant with respect to the image size, as can be seen in Fig.~\ref{fig:LRvsHR}.
In fact, the only operation in the low-resolution stream whose
runtime depends on the image size is the bilinear downsampling. Indeed,
$\LearnedDownsamplingFactor$ is kept fixed while $\DownsamplingFactor$ is adapted to the image size, so that the convolutional network's output resolution is always  $16 \times 16$ for inputs with a $1:1$ aspect ratio, regardless of the input image size. 

\myparagraph{Qualitative comparisons.}\label{sec:qualitative}
Examples of our translated images are shown in Fig.~\ref{fig:qualitative} and many more results are provided in the \href{https://tamarott.github.io/ASAPNet_web/resources/SM.pdf}{supplemental material (SM)}. 
Our method generates plausible outputs with high fidelity details, whose visual quality is at least comparable to the baselines (and even significantly better in some cases).

\myparagraph{Human perceptual study.}\label{sec:user_study}
To further assess the realism of the generated images, we conduct a user study on Amazon Mechanical Turk (AMT).
We adopt the protocol of~\cite{zhang2016colorful} and compare our models with CC-FPSE~\cite{liu2019learning}, SPADE~\cite{Park_2019_CVPR} and pix2pixHD~\cite{wang2018pix2pixHD}, which are the competitors with the best runtime-accuracy tradeoff (see Fig.~\ref{fig:eval}).
For each test (11 in total) we asked 50 users which
image looks more realistic in a Two Alternative Forced Choice (2AFC) test between our method and each baseline.
For each study we used 50 different pairs of images.
The results are summarized in Fig.~\ref{fig:amt}.
Across most resolutions, datasets and baselines, users rank our method at least on par with the baselines. User's ranking scores point that up to a $10\times$ speedup, ASAP-net does not incur any degradation in visual quality compared to baselines. At $18\times$ speedup, we experience a small degradation with respect to CC-FPSE on Cityscapes, but still maintain a significant advantage over CC-FPSE on Facades. 

\myparagraph{Quantitative evaluation.}\label{sec:quantitative}
We follow previous works~\cite{Park_2019_CVPR, wang2018pix2pixHD} and quantify the
quality of the generated images using:
\begin{inparaenum}[(i)]
\item the Fr\'echet Inception Distance (FID)~\cite{heusel2017gans}, which
   measures the similarity between the distributions of real and generated
   images, and 
\item semantic segmentation scores; for this we run the generated images trough a semantic segmentation network~\cite{yu2017dilated} (we use the \emph{drn-d-105} model for Cityscapes and the \emph{drn-d-22} for Facades) and measure how well they match the real segmentation map according to pixelwise accuracy and mean Intersection-over-Union (mIoU).
\end{inparaenum}
These evaluations are summarized in Fig.~\ref{fig:eval}.
As seen in the first row of Fig.~\ref{fig:eval}, our model achieves the best FID scores at the low runtime regime, populating a new region in the FID-runtime plane. 
%pix2pixHD~\cite{wang2020cnn}, SPADE~\cite{Park_2019_CVPR} and CC-FPSE~\cite{liu2019learning} 
This is true for all datasets and resolutions. Particularly, we obtain FID scores that are comparable to pix2pixHD and SPADE and significantly
better than pix2pix~\cite{isola2017image}, while operating at a comparable runtime to pix2pix (note the logarithmic runtime scale). This shows the speedups we obtain typically come on the expense of minimal degradation in visual quality, if any. 
A similar behavior is seen in terms of pixelwise accuracy and Mean IoU scores.
%

%
%Mean IoU scores are reported in Table~\ref{tab:mIoU}.
%
To further investigate FID score trends, in Fig.~\ref{fig:fid_curve} we compare our method with several variants of SPADE and pix2pixHD. When gradually cutting down their number of channels to reach the same runtime as ASAP-net, both these methods incur a significant degradation in performance.
%\myparagraph{FID vs. run-time curves.}
%\tamar{where? as part of model ablation? FID discussion?}

\subsection{Analysis}
We investigate the different properties of ASAP-Net.

\myparagraph{Beyond labels.} We qualitatively test the ability of ASAP-Net to perform additional tasks, other than translating labels to images. We train our model on the NYU dataset~\cite{silberman2012indoor} to predict depth maps from RGB images. As can be seen in Fig.~\ref{Fig:depth}, SPADE~\cite{Park_2019_CVPR} completely fails in this task, which involves continuous-valued inputs. This is because its normalization is specifically designed for discrete label maps. Our model, on the other hand, is not restricted to labels, and manages to learn a plausible depth-to-RGB mapping. %Please see additional results on SM.

\myparagraph{Model ablations.}\label{sec:ablation}
In Fig.~\ref{fig:ablation} we provide an ablation study illustrating the role of each design component.
Both the positional encoding~(Fig.~\ref{fig:abl:pos_encoding}) and the
spatially-varying functions~(Fig.~\ref{fig:abl:constant}) are critical for obtaining
realistic synthesis. 
We also quantify the impact of the downsampling factor on speed in
Figs.~\ref{fig:abl:speed1},~\ref{fig:abl:speed2}.

\begin{figure}
    %\centering
    %\begin{subfigure}[]{0.5\textwidth}
    %\centering
    %\includegraphics[width=1\linewidth]{figs/miou/miou_vs_runtime_v2_log.pdf}
    %\caption{mIoU}
    %\label{tab:mIoU:totmIoU}
    %\end{subfigure}
    %\\ \vspace{-0.5cm}
    %\begin{subfigure}[]{0.5\textwidth}
    \centering
    \includegraphics[width=1\linewidth]{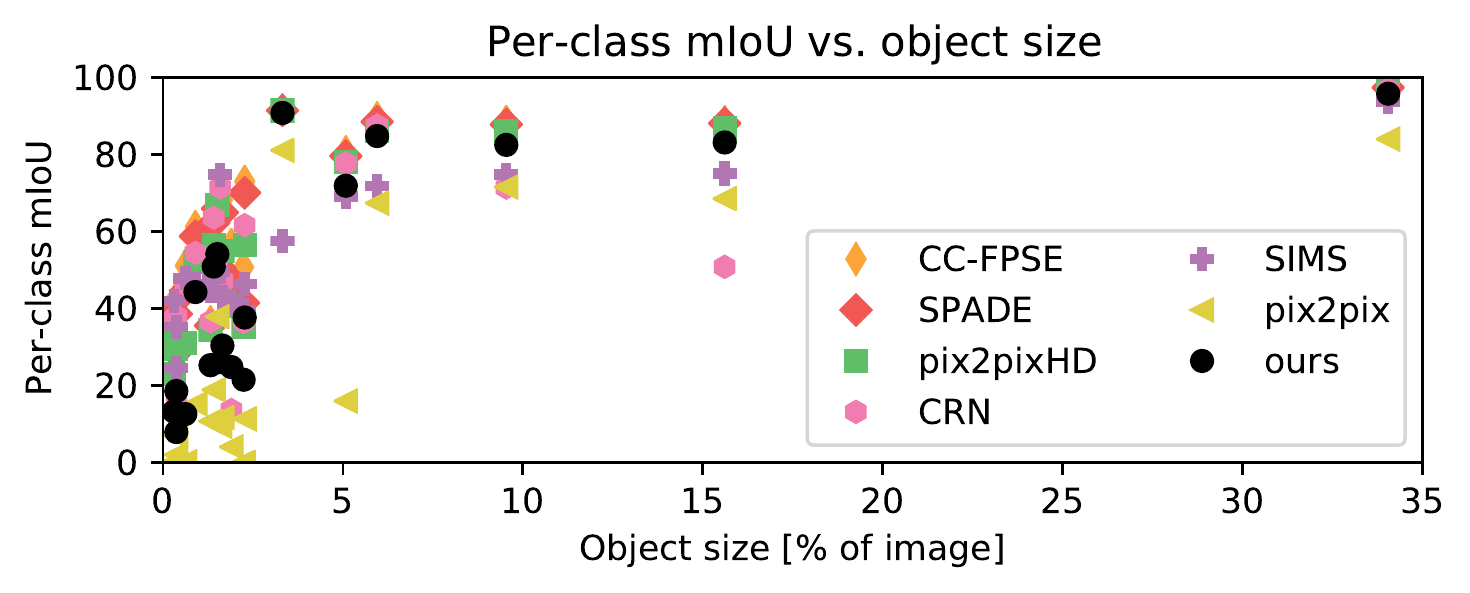} %\caption{Per-class mIoU}
    %\end{subfigure}
    \vspace{-0.7cm}
    \caption{\textbf{Per-class mIoU.}
    Our method sometimes misrepresents very small objects because of its extremely low resolution computation. When separating the mIoU scores (cityscapes $256\times512$) into classes, we can see a degradation for objects that are smaller than 3\% of the image area.
    %
    %Still, we outperform pix2pix by a large margin and are competitive on larger objects. %\todo{[add a scatter plot with avg. segment size vs. IoU. How mIoU is changed when working with larger image resolution at low-res stream?]}
    }
    \label{fig:mIoU}
    
    %\begin{tabular}{lcc}
    %   \toprule
    %   \multirow{2}{*}{method} & Cityscapes & Cityscapes \\
    %    & $256\times 512$ & $512 \times 1024$  
    %   \\
    %   \midrule
    %   CC-FPSE \cite{liu2019learning} & 65.5 & 71.0
    %   \\
    %   SPADE \cite{Park_2019_CVPR} & 62.1 & 73.6
    %   \\
    %   pix2pixHD \cite{wang2018pix2pixHD} & 57.1 & 60.8\\
    %   CRN~\cite{chen2017photographic} & 52.4 & 
    %   \\
    %   SIMS~\cite{qi2018semi} & 47.2 & 52.0
    %   \\
    %   pix2pix \cite{isola2017image} & 27.3 & 27.2 
    %   \\
    %   ASAP-Net (ours) & 46.1 & 42.8
    %   \\
    %   \bottomrule
    %\end{tabular} 
\end{figure}

\myparagraph{Limitations.} \label{sec:limitation}
Because most of our computations are done at a very low resolution (\eg
$\TotalDownsamplingFactor=64$), the accuracy of our method is limited for very
small object classes. %This does not severely affect the segmentation pixel
%accuracy (Fig.\ref{fig:eval}, second row). However
This is reflected in mIoU segmentation
scores (Fig.~\ref{fig:eval}, third row). Here, in some cases our scores are slightly inferior to 
those of the baselines, %CC-FPSE, SPADE and pix2pixHD, 
although usually %similar to CRN and SIMS and 
significantly higher than pix2pix.
When separating the scores into Per-class mIoU for Cityscapes (Fig.~\ref{fig:mIoU}), it is clear that the gap %is largest 
is due to small objects, which occupy less than $3\%$ of the image size.%like ``traffic signs'' and ``motorcycles''.

\section{Conclusion}

We propose a novel neural architecture for generative adversarial image-to-image
translation problems that is an order of magnitude faster than previous work,
yet maintains the same high level visual quality.
We achieve %The key to 
this speedup by %is 
a new image synthesis approach%to image synthesis
: at full resolution we use a collection of spatially-varying lightweight networks that operates %synthesizes the output 
in a pixel-wise fashion. 
The parameters of these networks are predicted by a convolutional network that runs at a much lower resolution.
Despite the low-resolution processing and pixelwise operations, our model can generate high-frequency details. This is due to the positional encoding of the input pixels.

%\input{sections/6_broader}

    % \newpage

{\small
\bibliographystyle{ieee_fullname}
\bibliography{bib}
}

\end{document}